\documentclass{article}
\usepackage{iclr2027_conference,times}
\usepackage{amsmath,amsfonts,bm}

\def\eqref#1{equation~\ref{#1}}
\def\1{\bm{1}}

\DeclareMathAlphabet{\mathsfit}{\encodingdefault}{\sfdefault}{m}{sl}
\SetMathAlphabet{\mathsfit}{bold}{\encodingdefault}{\sfdefault}{bx}{n}

\usepackage[hidelinks]{hyperref}
\usepackage{url}
\usepackage{amsmath,amssymb}
\usepackage{graphicx}
\usepackage{enumitem}
\usepackage{algorithm}
\usepackage{algpseudocode}
\usepackage{float}
\usepackage{placeins}
\usepackage{booktabs}
\usepackage{multirow}
\usepackage{subcaption}
\usepackage{colortbl}
\usepackage{xcolor}
\usepackage{wrapfig}
\usepackage[most]{tcolorbox}
\usepackage{amsthm}

\newlength{\wraptablesep}
\newcommand{\systemname}{AlgoEvo}

\definecolor{qbg}{RGB}{235,247,249}
\definecolor{qframe}{RGB}{158,218,225}
\newtcolorbox{qbox}{%
  enhanced, colback=qbg, colframe=qframe,
  leftrule=2pt, rightrule=0pt, toprule=0pt, bottomrule=0pt,
  arc=0pt, boxsep=2pt, left=8pt, right=8pt, top=6pt, bottom=6pt,
  before skip=10pt, after skip=2pt
}
\theoremstyle{plain}
\newtheorem{question}{Question}

\newtcolorbox{unitbox}{%
  colback=blue!5, colframe=blue!40, width=\linewidth, breakable,
  left=1mm, right=1mm, top=1mm, bottom=1mm,
  before skip=4pt, after skip=4pt
}

\definecolor{hlrowbg}{RGB}{231,240,254}
\newcommand{\hlrow}[1]{%
  \begingroup\setlength{\fboxsep}{2pt}%
  \colorbox{hlrowbg}{#1}%
  \endgroup}

\definecolor{tabheadbg}{HTML}{D9E8F6}   % column-header rows (soft sky blue)
\definecolor{taboursbg}{HTML}{D7F0D7}   % our method's rows (soft mint)
\definecolor{grpA}{HTML}{D6E6F5}    % group headers (single / multi-obj / multi-comp)
\definecolor{grpB}{HTML}{E6D4F2}
\definecolor{grpC}{HTML}{FBE0BC}
\definecolor{hposA}{HTML}{81C784}   % our margin: strong / medium / mild
\definecolor{hposB}{HTML}{A5D6A7}
\definecolor{hposC}{HTML}{E8F5E9}
\definecolor{hneg}{HTML}{F7D9D9}    % our deficit
\newcommand{\hpA}[1]{\cellcolor{hposA}#1}
\newcommand{\hpB}[1]{\cellcolor{hposB}#1}
\newcommand{\hpC}[1]{\cellcolor{hposC}#1}
\newcommand{\hn}[1]{\cellcolor{hneg}#1}

\usepackage{listings}
\lstdefinestyle{boxed}{%
  basicstyle=\footnotesize\ttfamily,
  escapeinside={(*@}{@*)},
  columns=fullflexible, keepspaces=true, showstringspaces=false,
  breaklines=true, breakindent=2em, breakautoindent=true,
  frame=none, aboveskip=0pt, belowskip=0pt, xleftmargin=0pt
}

\definecolor{promptframe}{RGB}{0,80,160}
\definecolor{promptbg}{RGB}{240,248,255}
\newtcolorbox{prompttemplate}{%
  enhanced, colback=promptbg, colframe=promptframe,
  fonttitle=\bfseries\small, title={System Prompt Example},
  boxrule=0.8pt, arc=3pt, left=4pt, right=4pt, top=4pt, bottom=4pt,
  width=\linewidth, breakable,
  before skip=8pt, after skip=6pt,
  attach boxed title to top left={yshift=-2mm, xshift=4mm},
  boxed title style={colback=promptframe, colframe=promptframe, arc=2pt}
}
\definecolor{instrframe}{RGB}{0,110,70}
\definecolor{instrbg}{RGB}{240,252,246}
\newtcolorbox{instructiontemplate}{%
  enhanced, colback=instrbg, colframe=instrframe,
  fonttitle=\bfseries\small, title={Subtask Reference Instructions},
  boxrule=0.8pt, arc=3pt, left=4pt, right=4pt, top=4pt, bottom=4pt,
  width=\linewidth, breakable,
  before skip=8pt, after skip=6pt,
  attach boxed title to top left={yshift=-2mm, xshift=4mm},
  boxed title style={colback=instrframe, colframe=instrframe, arc=2pt}
}
\definecolor{codeframe}{RGB}{95,95,95}
\definecolor{codebg}{RGB}{249,249,249}
\newtcolorbox{codebox}{%
  enhanced, colback=codebg, colframe=codeframe,
  boxrule=0.6pt, arc=2pt, left=4pt, right=4pt, top=4pt, bottom=4pt,
  width=\linewidth,
  before skip=6pt, after skip=6pt
}

\newcommand{\papertitle}{\systemname{}: Self-Evolving Agentic Search for Automated Algorithm Discovery}
\title{\papertitle}

\author{Junhao Qiu\textsuperscript{1}, Qinglong Hu\textsuperscript{1}, Ji Cheng\textsuperscript{1}, Xialiang Tong\textsuperscript{2}, Liyong Lin\textsuperscript{3}, Qingfu Zhang\textsuperscript{1} \\[4pt]
\textsuperscript{1}Department of Computer Science, City University of Hong Kong\\
\textsuperscript{2}Huawei Noah's Ark Lab\\
\textsuperscript{3}Institute of Advanced Intelligence and Computing, A*STAR\\
\texttt{junhaoqiu2-c@cityu.edu.hk, qingfu.zhang@cityu.edu.hk} \\
}

\iclrfinalcopy  % arXiv version: show real authors
\begin{document}

\maketitle
% arXiv version: header on the right (no venue mention)
\lhead{}\rhead{Preprint}

% ---- arXiv version (A): affiliation logos below the author block ----
\vspace*{-2em}
\begin{center}
\raisebox{-0.5\height}{\includegraphics[height=1.25cm]{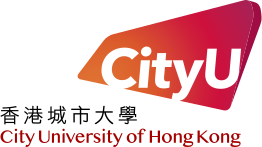}}%
\hspace{1.1em}\raisebox{-0.5\height}{\textcolor{gray!35}{\rule{0.4pt}{0.95cm}}}\hspace{1.1em}%
\raisebox{-0.5\height}{\includegraphics[height=0.58cm]{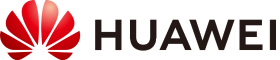}}%
\hspace{1.1em}\raisebox{-0.5\height}{\textcolor{gray!35}{\rule{0.4pt}{0.95cm}}}\hspace{1.1em}%
\raisebox{-0.5\height}{\includegraphics[height=0.9cm]{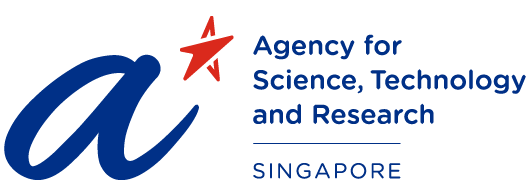}}
\vspace{0.3em}
\end{center}

% Drafting aid: suppresses the main text from the appendix-only contents list.
% The marker is written into the .toc file and takes effect from the next pass.
% Remove together with the appendix contents block before submission.
\addtocontents{toc}{\protect\setcounter{tocdepth}{-1}}

\begin{abstract}
Large language models have advanced automated algorithm discovery by synthesizing executable code, but existing frameworks trap them in rigid search pipelines with pre-defined control flows. This limitation restricts adaptive reasoning, blocks cross-paradigm transfer, and overlooks richer execution feedback. To bridge this gap, we introduce an end-to-end framework, \textbf{\systemname{}}, a unified agentic architecture that transforms automated algorithm discovery into an interactive, knowledge-accumulating process. An autonomous agent dynamically inspects, diagnoses, and edits code based on runtime feedback. A design skill hub decouples paradigm-specific knowledge from the core discovery engine, allowing a unified workflow to seamlessly handle single-heuristic, multi-objective, and multi-component design. Meanwhile, a hierarchical experience bank organizes search trajectories into a task-level tree to guide exploration and consolidates cross-task patterns into reusable skills. Across six representative benchmark tasks, \systemname{} reaches state-of-the-art performance with as little as 7\% of the evaluation budget and reduced token consumption, demonstrating strong intra-task accumulation, cross-task transfer, and the ability to reproduce or exceed the strongest existing methods through flexible skill activation.
\end{abstract}

% ============================================================
% INTRODUCTION (V1) - polished from V0, contributions from V1.md
% ============================================================
\section{Introduction}

Automated Algorithm Discovery (AAD) focuses on generating optimization heuristics without manual algorithm engineering~\citep{burke2013hyper, stutzle2019automated}. Integrating Large Language Models into Automated Heuristic Design (LLM-AHD) has enabled direct code synthesis for complex heuristics~\citep{llm4ad2026survey,wu2024evolutionary}. From the perspective of search mechanisms, existing methods have explored various mature evolutionary paradigms~\citep{eoh2024,reevo2024,llamea2025}. These include population-based global search, as in EoH~\citep{eoh2024}, island models that maintain high-quality diversity, as in FunSearch~\citep{funsearch2024}, Monte Carlo tree search (MCTS)~\cite{mctsahd2025}, and iterative search over localized code edits~\citep{zhang2024understanding}. As the design target has grown from a single scoring function into a complex multi-stage algorithmic topology, these mechanisms have been extended to meet increasingly diverse design requirements. This progression spans from single-objective optimization to Pareto trade-offs in multi-objective settings, as in MEoH~\citep{meoh2025}, and to multi-component operator systems, as in E2OC~\citep{e2oc2026} and MOTIF~\citep{motif2025}, achieving strong empirical performance across a variety of optimization and scheduling tasks.

\begin{figure}[t]
\centering
\includegraphics[width=\linewidth]{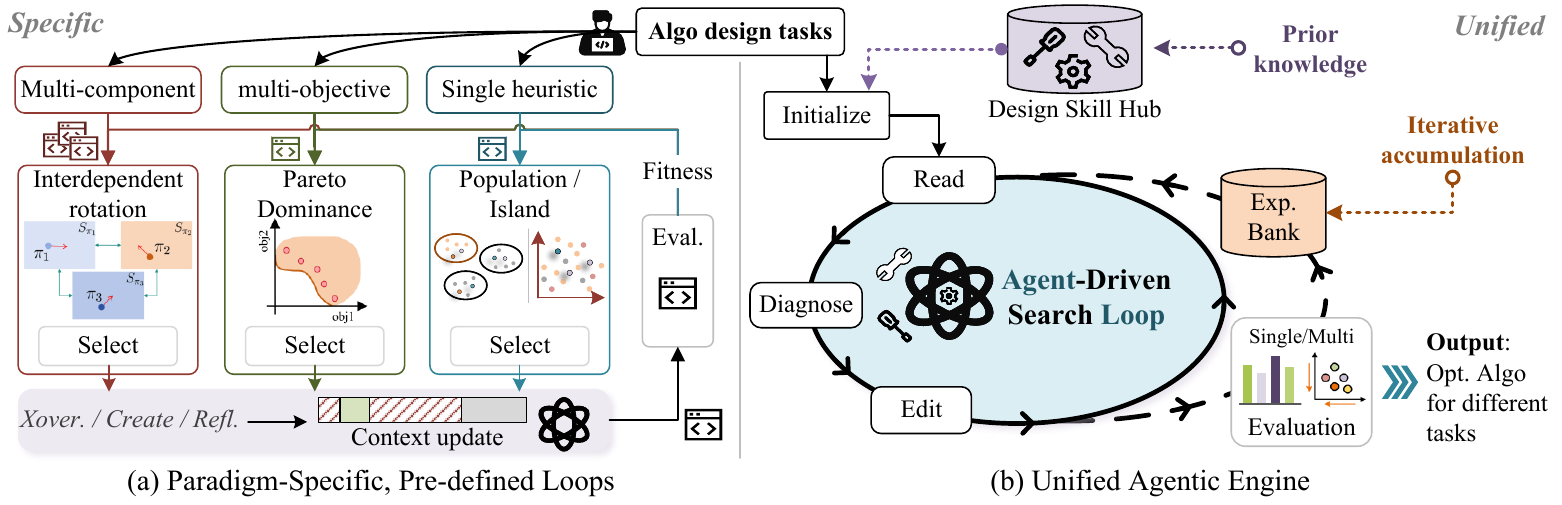}
\caption{Comparison of search paradigms in LLM-driven AAD. (a) Conventional methods rely on rigid, task-specific pipelines with pre-defined loops and passive LLM sampling, leading to framework fragmentation. (b) \systemname{} decouples the search loop via a design skill hub and experience bank, enabling an autonomous agent to handle diverse optimization paradigms through interactive tool use and iterative knowledge accumulation.
}
\label{fig:teaser}
\end{figure}

Despite these growing capabilities, current frameworks encounter limitations rooted in their control flow, modularity, and memory management, as illustrated in Figure~\ref{fig:teaser}a. First, they typically embed large language models within rigid, \textbf{expert-predefined evolutionary pipelines} rather than empowering them as active search orchestrators. In this setup, the model often functions as a passive code sampler driven by static prompt templates, which restricts interactive debugging, diagnostic analysis, and deep structural refactoring~\citep{wang2024promptagent}. Second, \textbf{paradigm-specific coupling} causes severe framework fragmentation, requiring practitioners to maintain separate codebases for distinct design tasks. Third, because these pipelines evaluate candidates using scalar performance feedback~\citep{reevo2024,qiu2026evodr}, \textbf{amnesic trial-and-error search} often discards the broader structured execution details required to understand why a heuristic succeeded or failed. Consequently, many discovery workflows risk reducing each problem to an isolated trial-and-error procedure that struggles to accumulate or transfer reusable algorithmic experience.

To address these architectural and memory constraints, we present \textbf{\systemname{}}, an agentic framework that replaces predefined prompt completion with an interactive, self-evolving discovery loop. Rather than following a rigid schedule, \systemname{} empowers an autonomous agent to direct the discovery process, dynamically determining when to inspect code, run diagnostic analyses, edit components, or evaluate candidates. Operating at the core of this flexible loop, the design skill hub expands the boundaries of automated design by decoupling paradigm-specific knowledge from the search mechanism, allowing a unified agentic process to seamlessly span single-heuristic, multi-objective, and multi-component tasks without structural modification. Concurrently, the hierarchical experience bank deepens optimization within individual tasks and across generations by transforming historical search trajectories into structured experience cards, organizing them into a task-level tree, and retrieving context-relevant insights to guide subsequent iterations. Through this synergy, the framework elevates automated algorithm discovery from isolated trial-and-error routines into a cumulative, self-improving scientific process (Figure~\ref{fig:teaser}b).

Our main contributions can be summarized as follows:
\begin{itemize}[leftmargin=*,itemsep=2pt,topsep=2pt]

\item We propose \textbf{\systemname{}}, an end-to-end framework for automated algorithm discovery that integrates an agentic search loop with the design skill hub. Instead of embedding a large language model within a predefined evolutionary pipeline, \systemname{} delegates the discovery process to an autonomous agent that dynamically reads, edits, diagnoses, and evaluates algorithmic code across single-heuristic, multi-objective, and multi-component settings.

\item We develop a hierarchical experience bank that transforms historical search trajectories into structured experience cards and organizes them through a task-level tree for context-aware retrieval. It accumulates and reuses algorithmic experience, ensuring that prior knowledge continually contributes to the evolution and transfer of solutions across discovery processes.

\item We conduct extensive experiments across single-heuristic, multi-objective, and multi-component algorithm-design tasks, comparing against specialized methods such as FunSearch, EoH, MEoH, E2OC, and MOTIF. \systemname{} reaches state-of-the-art quality within $7\%$ of the evaluation budget and matches or surpasses the strongest baseline on all six tasks, by up to $8.8\%$, and further studies verify continual self-improvement and cross-task knowledge transfer.
\end{itemize}

\section{Automated Algorithm Discovery}
\label{sec:aad}

Automated algorithm discovery seeks to synthesize executable code or heuristic programs for target optimization problems~\citep{llm4ad2026survey,hu2026partition}, ranging from single heuristics to multi-objective strategies and interacting operators for complex systems. The discovery process navigates a candidate program space guided by task-specific evaluations; we formalize this setting as a discovery task $\mathcal{T}$.

\paragraph{Definition 2.1 (Domain and Instance).}
An optimization domain $\mathcal{D}$ is characterized by an instance space $\mathcal{X}_d$, a solution space $\mathcal{Y}_d$, and a task objective $f_d:\mathcal{X}_d\times\mathcal{Y}_d\rightarrow\mathbb{R}$. For an instance $\mathbf{x}\in\mathcal{X}_d$ with feasible set $\mathcal{Y}_d(\mathbf{x})\subseteq\mathcal{Y}_d$, the underlying optimization problem is
\begin{equation}
\mathbf{y}^{*}
=
\arg\min_{\mathbf{y}\in\mathcal{Y}_d(\mathbf{x})}
f_d(\mathbf{x},\mathbf{y}).
\end{equation}

\paragraph{Definition 2.2 (Solver and Algorithm).}
A solver $s$ generates a solution using a collection of designable strategies $\mathbf{\Pi}=(\pi_1,\ldots,\pi_K)$ with $\pi_k\in\mathcal{S}_k$, where $\mathcal{S}_k$ is the search space of the $k$-th strategy. Each strategy may represent a scoring rule, construction policy, neighborhood operator, penalty update mechanism, or another component. The induced algorithm space is $\mathcal{S}=\mathcal{S}_1\times\cdots\times\mathcal{S}_K$, where $K=1$ corresponds to single-heuristic design and $K>1$ to multi-component design with strategies optimized jointly. The solver's output on instance $\mathbf{x}$ is $s(\mathbf{x}\mid\mathbf{\Pi})$.

\paragraph{Definition 2.3 (Algorithm Optimization).}
The performance of an algorithm $\mathbf{\Pi}$ on instance $\mathbf{x}$ is measured by
\begin{equation}
F_d(\mathbf{x}\mid\mathbf{\Pi})
=
\phi_d
\left(
f_d\left(
\mathbf{x},
s(\mathbf{x}\mid\mathbf{\Pi})
\right)
\right),
\label{eq:solver-performance}
\end{equation}
where $\phi_d$ maps the task-specific evaluation to a scalar measure, i.e., the objective value for single-objective tasks or a scalar indicator such as hypervolume (HV) for multi-objective tasks. We normalize the direction so that smaller $F_d$ is better, and define the discovery objective as
\begin{equation}
\mathbf{\Pi}^{*}
=
\arg\min_{\mathbf{\Pi}\in\mathcal{S}}
\mathbb{E}_{\mathbf{x}\sim\mathcal{X}_d}
\left[
F_d(\mathbf{x}\mid\mathbf{\Pi})
\right],
\label{eq:design-objective}
\end{equation}
subject to a computational budget $B$ bounding evaluation and discovery resources.

\section{\systemname{}: Self-Evolving Agentic Algorithm Discovery}
\label{sec:algoevo}

\begin{figure*}[t]
\centering
\includegraphics[width=\textwidth]{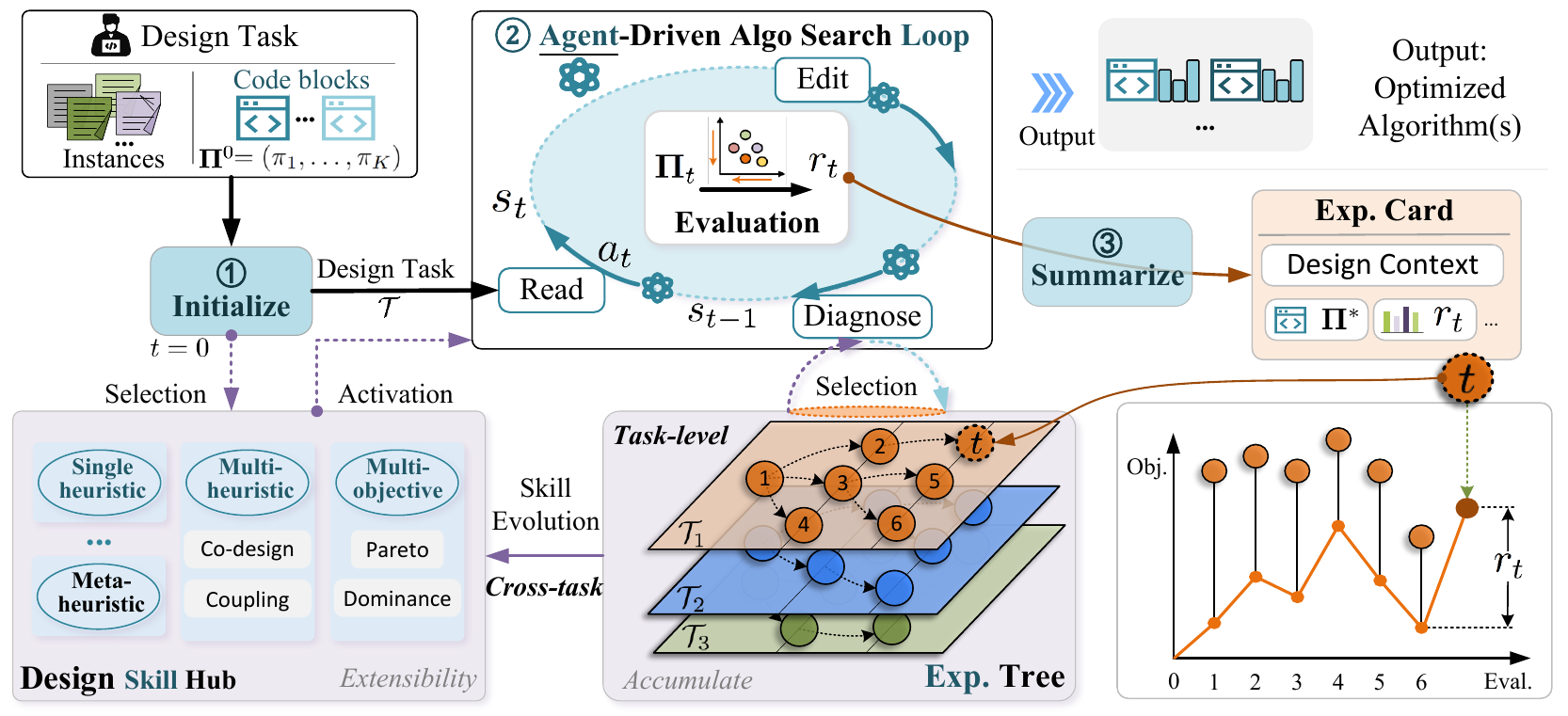}
\caption{Overview of \systemname{}. Given a target design task, (1) the design skill hub activates the appropriate algorithm design paradigm; (2) an autonomous agent drives the discovery process through an interactive search loop of code inspection, diagnosis, editing, and evaluation; and (3) performance feedback generates structured experience cards. These cards populate a task-level experience tree that guides ongoing exploration, while successful patterns accumulate across tasks to continually evolve the skill hub.}

	\label{fig:framework}
\end{figure*}

\systemname{} formulates automated algorithm design as an agentic discovery process over the algorithm space $\mathcal{S}$. Unlike conventional pipelines with fixed generation-and-evaluation cycles, an autonomous agent dynamically inspects, modifies, evaluates, and diagnoses algorithms based on runtime feedback (Figure~\ref{fig:framework}). This process is supported by two complementary components: the design skill hub, which provides paradigm-specific execution contracts to let a single loop span single-heuristic, multi-objective, and multi-component tasks; and the hierarchical experience bank, which stores distilled trajectories and retrieves context-relevant insights to guide exploration. Concretely, the pipeline proceeds in five stages: (1) a design skill is \emph{activated} for the task; (2) an experience tree is \emph{matched} or created; (3) the agent \emph{searches} by inspecting, editing, diagnosing, and evaluating; (4) each evaluation is \emph{summarized} into an experience card; and (5) across tasks, accumulated experience \emph{refines} the skill. Through this synergy, flexible agentic actions and experience-driven selection drive both intra-task optimization and continual cross-task self-evolution.

\subsection{Agentic Search Loop}
\label{sec:loop}

\systemname{} explores the algorithm space through an \emph{agent-driven loop}. Unlike rigid pipelines with predefined generation-evaluation cycles, the agent dynamically decides its execution trajectory, observing the discovery state, selecting and executing an operation, and processing the feedback until the budget is exhausted or it terminates.

At step $t$, the discovery state is $s_t=(\mathbf{\Pi}_t,\mathcal{C}_t,\epsilon,F_d(\mathbf{x}\mid\mathbf{\Pi}_t))$, comprising the current algorithm $\mathbf{\Pi}_t$, working context $\mathcal{C}_t$, experience tree $\epsilon$, and performance feedback $F_d(\cdot)$. The working context maintains code, recent modifications, diagnostic observations, and intermediate reasoning. 

Given $s_{t-1}$, the agent selects an action $a_t$ to inspect code, modify a strategy, run a diagnosis, or evaluate a candidate, thus determining both the operation and its timing. Executing $a_t$ yields a transition $\tau_t=(s_{t-1},a_t,r_t,s_t)$. When $a_t$ updates the algorithm from $\mathbf{\Pi}_{t-1}$ to $\mathbf{\Pi}_t$, the immediate reward is the performance improvement, positive when the update lowers $F_d$,
\begin{equation}
r_t
=
F_d(\mathbf{x}\mid\mathbf{\Pi}_{t-1})
-
F_d(\mathbf{x}\mid\mathbf{\Pi}_t),
\label{eq:reward}
\end{equation}
so that maximizing the cumulative improvement $\sum_{t=1}^{T} r_t$ reduces to minimizing the final performance $F_d(\mathbf{x}\mid\mathbf{\Pi}_T)$, thereby recovering the global objective in Eq.~\ref{eq:design-objective}. Rather than following a fixed cycle, the agent interleaves diagnosis, modification, and evaluation, invoking targeted analysis when bottlenecks arise and structural edits when supported by evidence.

\subsection{Design Skill Hub}
\label{sec:hub}

The algorithm space $\mathcal{S}$ specifies what can be designed, but its structure is paradigm-dependent. A single-heuristic constructive rule, a multi-objective evolutionary operator, and a coupled destroy-repair system differ in the roles their strategies play, the interfaces they expose, and the principles that make them effective. Encoding these differences into the search procedure would yield a separate pipeline for each paradigm, reproducing the fragmentation we seek to avoid. \systemname{} therefore externalizes them as pluggable \emph{design skills} in a shared design skill hub $\mathcal{S}_{\mathrm{hub}}$, so that a new paradigm is supported by adapting the same general engine with an additional skill rather than by modifying it.

A design skill declares (i) the roles of the designable strategies in $\mathbf{\Pi}=(\pi_1,\ldots,\pi_K)$, (ii) their code interfaces, (iii) the principles that guide valid modifications, and (iv) the task-specific evaluation conventions. Together, these fix the constraints and inductive bias of the search over $\mathcal{S}$, while the search mechanism remains unchanged. Each skill follows a life cycle, being selected and activated when a task begins and refined once the task completes.

\paragraph{Selection and activation ($t=0$).}
At task initialization, the hub matches the task to the skill whose declared conditions it satisfies and activates it in the agent's context. The agent then uses the skill to interpret the design roles, reason about valid modifications, and construct algorithms within the corresponding strategy spaces $\mathcal{S}_1,\ldots,\mathcal{S}_K$. Because the paradigm is carried by the skill rather than the loop, the same engine supports all three design settings by changing the skill alone; the skill is thus what specializes a general agent to a particular design paradigm.

\paragraph{Evolution ($t=T$).}
At task completion, the evidence accumulated during discovery is consolidated into the experience bank; the skill is not updated within a single task, so evolution is triggered only across tasks. Each skill defines the paradigm under which experiences are accumulated, so that retrieved experiences are interpreted within the corresponding design context. Once a skill owns at least $\mu$ experience trees, the agent summarizes their most salient patterns and refines the skill. The hub therefore evolves with the tasks it supports, so that later tasks start from a stronger prior. A skill may also encode methodological knowledge from existing algorithm-design approaches, inheriting their inductive bias while retaining the autonomous execution process.

\subsection{Hierarchical Experience Bank}
\label{sec:expbank}

The design experiences produced by the agentic loop are organized in a hierarchical experience bank $\mathcal{E}$ with three levels: experience cards, a task-level experience tree, and cross-task consolidation.

\paragraph{Level 1: Experience cards.}
The agent's actions differ in whether they yield a performance signal. Evaluation produces the reward $r_t$ of Eq.~\ref{eq:reward}, while actions that inspect, edit, diagnose, reason, or reflect change the algorithm or context without producing one. The bank is therefore updated at evaluation events, each distilled into an \textbf{experience card} recording the design context, the modification, the reward, and the rationale. Cards are the basic units of the bank and persist across evaluations and tasks.

\paragraph{Level 2: Experience tree.}
Cards from the same task form an experience tree $\epsilon$, where nodes are cards and edges represent derived-from relationships. The tree is traversed using the four operations of MCTS~\citep{swiechowski2023monte}.

\emph{Selection.} A card is selected to guide the next design decision. Each card $e$ maintains a mean reward $\bar{r}(e\mid q)$ and a visit count $n(e,q)$ under the current situation $q$, and selection maximizes the upper confidence bound
\begin{equation}
\mathrm{UCB}(e\mid q)
=
\bar{r}(e\mid q)
+
c\sqrt{\frac{2\ln N}{n(e,q)}},
\label{eq:ucb}
\end{equation}
where $N$ is the tree-level visit statistic and $c$ controls the exploration-exploitation trade-off. The situation $q$, written $q_t$ at step $t$, is the set of conditions active in the current search state, drawn from $\mathcal{Q}=\{\emph{stagnation},\emph{bottleneck},\emph{coupling},\emph{sparse front}\}$ and detected by the diagnostic engine from runtime attribution, interaction, and search-history signals (Appendix~\ref{app:method-diag}). Since several conditions may hold at once, a card is scored by the statistics it has accumulated under those it matches. When $n(e,q)=0$, the global mean of $e$ serves as a prior with an exploration incentive.

\emph{Expansion.} When a design decision yields a verified candidate, its card is attached as a child of the card it evolved from.

\emph{Evaluation.} The candidate is evaluated, producing the reward $r_t$ of Eq.~\ref{eq:reward}.

\emph{Backpropagation.} The reward is propagated to the ancestors of the new card $c_t$, so that a design decision is credited for the outcomes of the candidates it produced. Each ancestor $e$ updates its mean reward $\bar{r}(e\mid q)$ and visit count $n(e,q)$ under the current situation $q$, so that design directions that consistently produced strong descendants are preferred in subsequent selections.

\paragraph{Level 3: Cross-task consolidation.}
The experience tree captures knowledge within a single task; across tasks, the bank accumulates broader evidence. Once a skill owns at least $\mu$ experience trees, the agent summarizes their most salient patterns and refines the skill, transferring concrete experience into paradigm-level knowledge.

\begin{algorithm}[t]
\caption{Agentic Algorithm Discovery.}
\label{alg:algoevo}
\begin{algorithmic}[1]
\Require Design task $\mathcal{T}$, skill hub $\mathcal{S}_{\mathrm{hub}}$, experience bank $\mathcal{E}$, budget $B$, threshold $\mu$
\Ensure Optimized algorithm $\mathbf{\Pi}^{*}$
\State $\sigma \leftarrow \mathrm{\textbf{Activate}}(\mathcal{T}, \mathcal{S}_{\mathrm{hub}})$ \Comment{1. Activate skill}
\State $\epsilon \leftarrow \mathrm{\textbf{Match}}(\mathcal{T}, \sigma, \mathcal{E})$ \Comment{2. Match/create tree (Algo. ~\ref{alg:init})}
\State Initialize $\mathbf{\Pi} \leftarrow \mathbf{\Pi}_0$, $\mathcal{C}_t$, $t \leftarrow 0$
\Repeat
    \State $q_t \leftarrow \mathrm{\textbf{Categorize}}(s_t)$ \Comment{3. Detect situation (Algo. ~\ref{alg:app-situation})}
    \State $a_t \leftarrow \mathrm{\textbf{Agent}}(\sigma, q_t, \epsilon, \mathcal{C}_t)$; execute $a_t$; update $\mathcal{C}_t$ \Comment{Inspect/edit/diagnose/retrieve/evaluate}
    \State \textbf{if} $a_t = \text{evaluate}$: update $\mathbf{\Pi}$ and $r_t$, summarize card to $\epsilon$ \Comment{4. Summarize (Algo. ~\ref{alg:distill})}
    \State \textbf{if} $a_t \in \{\text{diagnose}, \text{retrieve}\}$: select card $e$ via Eq.~\ref{eq:ucb} \Comment{Use card}
    \State $t \leftarrow t + 1$
\Until{$B$ exhausted or $a_t = \text{terminate}$}
\State $\mathrm{\textbf{Evolve}}(\sigma, \mathcal{E}, \mathcal{S}_{\mathrm{hub}}, \mu)$ \Comment{5. Refine skill (Algo. ~\ref{alg:skill-evolution})}
\State \Return $\arg\min_{\mathbf{\Pi}} F_d(\mathbf{x}\mid\mathbf{\Pi})$
\end{algorithmic}
\end{algorithm}

\subsection{Experience Accumulation and Self-Evolution}
\label{sec:evolution}

The self-evolution of \systemname{} follows a closed-loop MDP process (Algorithm~\ref{alg:algoevo}). At initialization (Algorithm~\ref{alg:init}), the system generates a system prompt from the problem description, then uses this prompt to match a skill and retrieve or create an experience tree. Once initialized, the agent autonomously drives the discovery process, selecting at each decision step $t$ an action $a_t$ based on the current state and updating the working context with the outcome. When $a_t$ is \emph{evaluate}, a new experience card is created and attached to the tree; when $a_t$ is \emph{diagnose} or \emph{retrieve}, the agent retrieves relevant experiences to guide its reasoning.

Beyond single-task adaptation, the framework achieves cumulative evolution via cross-task knowledge consolidation (Algorithm~\ref{alg:skill-evolution}). Once a skill owns at least $\mu$ experience trees, the agent summarizes their most salient patterns and refines the skill accordingly. This hierarchical propagation allows the overarching discovery engine to continuously enhance its foundational capabilities and transfer expertise to new domains without requiring manual redesign or structural modifications.

\section{Experiments}
\label{sec:experiments}

\subsection{Experimental Setup}
\label{sec:exp-setup}

\paragraph{Benchmark.}
The benchmark comprises six problems across three paradigms, with objectives, instance sets, and code contracts detailed in Appendix~\ref{app:tasks}: constructive heuristic design for TSP and CVRP, multi-objective design for Bi-TSP and Bi-FJSP, and multi-component co-design for CVRP-DR and FJSP 4-Ops~\citep{eohs2026,chen2023neural,motif2025,brandimarte1993fjsp}. Each task features disjoint training and held-out test sets, with all instances, random seeds, and evaluation scripts publicly released.

\paragraph{Baselines and protocol.}
We compare against each paradigm's strongest method. On TSP and CVRP these are nearest-neighbor heuristics and the LLM-AHD methods EoH~\citep{eoh2024}, ReEvo~\citep{reevo2024}, MCTS-AHD~\citep{mctsahd2025}, and FunSearch~\citep{funsearch2024}; on the multi-objective tasks, MEoH~\citep{meoh2025}, NSGA-II~\citep{deb2002nsga2}, and MOEA/D~\citep{zhang2007moead}; and on the multi-component tasks, MOTIF~\citep{motif2025}, E2OC~\citep{e2oc2026}, and the EoH variants Synergy and Rotating-EoH. All methods run on the LLM4AD platform~\citep{liu2024llm4ad} under one shared protocol, with identical instance splits and evaluation entries, three seeds, and $500$ evaluations per seed, using \texttt{gpt-4o-mini} on TSP and CVRP and \texttt{DeepSeek-V4-Flash} elsewhere. Because \systemname{} writes and runs its own code, its budget is guarded against evasion by cross-checked counters, and any run that scores a candidate outside the shared evaluation entry is discarded rather than reported (Appendix~\ref{app:anti-cheat}). Remaining parameters are in Appendix~\ref{app:baselines}.

\paragraph{Metrics.}
For multi-objective results, a shared global normalization basis ensures comparability of HV and Inverted Generational Distance (IGD) within each problem. Appendix~\ref{app:metrics} formalizes these metrics and notes that multi-objective rows report training-instance quality since the final algorithm is fixed post-run. In Table~\ref{tab:main_results}, the evaluation column reports the mean number of candidate evaluations consumed per design task, and the token column reports cumulative backbone usage in millions. All reported dispersions denote sample standard deviations over random seeds.
% \begin{table}[h]
% \centering
% \caption{Benchmark.}
% \label{tab:benchmarks}
% \begin{tabular}{@{}llll@{}}
% \toprule
% Paradigm & Problem & Objective & Train / Test \\
% \midrule
% \multirow{2}{*}{Single heuristic} & TSP & tour length & 16$\times$50 / 24 + 6 TSPLib \\
%  & CVRP & route cost & 16$\times$50 / 16$\times$(50/100/200) \\
% \midrule
% \multirow{2}{*}{Multi-objective} & Bi-TSP & two tour lengths & 1$\times$50 \\
%  & Bi-FJSP & makespan, utilization & mk01--mk10 \\
% \midrule
% \multirow{2}{*}{Multi-component} & CVRP-DR & route cost & 10$\times$50 / 64 \\
%  & FJSP 4-Ops & makespan & mk01--mk10 / mk11--mk15 \\
% \bottomrule
% \end{tabular}
% \end{table}
\begin{table}[!b]
\centering
\footnotesize
\setlength{\tabcolsep}{4pt}
\caption{Performance on all six tasks. On the quality metrics the best result is \textbf{bold} and the runner-up \underline{underlined}; on the Evaluations and Tokens columns, where smaller is better, the best result is \textbf{bold}. HV is the only metric for which larger is better. The columns are defined in Section~\ref{sec:exp-setup}.}
\label{tab:main_results}
\begin{tabular}{@{}llcclc@{}}
\toprule
\cellcolor{grpA}\textbf{Single heuristic} & \textbf{Method} & \textbf{Train} & \textbf{Test} & \textbf{Evals} & \textbf{Tokens (M)} \\
\midrule
\multirow{6}{*}{TSP} 
 & NN & $6.824$ & $7.990$ & -- & -- \\
 & EoH & $\underline{6.309 \pm 0.043}$ & $7.392 \pm 0.058$ & $500$ & $4.1$ \\
 & ReEvo & $6.453 \pm 0.097$ & $7.699 \pm 0.220$ & $500$ & $4.3$ \\
 & MCTS-AHD & $6.331 \pm 0.040$ & $\underline{7.337 \pm 0.028}$ & $500$ & $3.8$ \\
 & FunSearch & $6.465 \pm 0.023$ & $7.484 \pm 0.051$ & $500$ & $3.2$ \\
 & \systemname{} & \hpB{$5.986 \pm 0.11$} & \hpC{$7.007 \pm 0.04$} & \hpA{$39$} & \hpB{$2.9$} \\
\midrule
\multirow{5}{*}{CVRP} 
 & NN & $13.611$ & $26.283$ & -- & -- \\
 & EoH & $13.537 \pm 0.028$ & $26.148 \pm 0.169$ & $500$ & $4.1$ \\
 & ReEvo & $13.403 \pm 0.129$ & $\underline{25.950 \pm 0.517}$ & $500$ & $4.3$ \\
 & MCTS-AHD & $\underline{13.236 \pm 0.302}$ & $25.967 \pm 0.447$ & $500$ & $4.1$ \\
 & FunSearch & $13.584 \pm 0.030$ & $26.270 \pm 0.013$ & $500$ & $3.9$ \\
 & \systemname{} & \hpC{$12.698 \pm 0.39$} & \hpB{$24.150 \pm 0.40$} & \hpA{$35$} & \hpA{$1.9$} \\
\midrule
\cellcolor{grpB}\textbf{Multi-objective} & \textbf{Method} & \textbf{HV} $\uparrow$ & \textbf{IGD} $\downarrow$ & \textbf{Evals} & \textbf{Tokens (M)} \\
\midrule
\multirow{4}{*}{Bi-TSP} 
 & MEoH & $\underline{0.760 \pm 0.020}$ & $\underline{0.077 \pm 0.018}$ & $500$ & $4.2$ \\
 & NSGA-II & $0.614 \pm 0.067$ & $0.215 \pm 0.067$ & $500$ & $3.6$ \\
 & MOEA/D & $0.314 \pm 0.180$ & $0.601 \pm 0.172$ & $500$ & $3.5$ \\
 & \systemname{} & \hpB{$0.827 \pm 0.012$} & \hpA{$0.036 \pm 0.003$} & \hpA{$36$} & \hpA{$2.6$} \\
\midrule
\multirow{4}{*}{Bi-FJSP} 
 & MEoH & $\underline{0.903 \pm 0.051}$ & $\mathbf{0.143 \pm 0.077}$ & $500$ & $6.3$ \\
 & NSGA-II & $0.899 \pm 0.048$ & $\underline{0.149 \pm 0.067}$ & $500$ & $5.7$ \\
 & MOEA/D & $0.792 \pm 0.002$ & $0.286 \pm 0.000$ & $500$ & $5.2$ \\
 & \systemname{} & \hpC{$0.911 \pm 0.083$} & \hn{$0.162 \pm 0.125$} & \hpA{$33$} & \hpA{$2.4$} \\
\midrule
\cellcolor{grpC}\textbf{Multi-component} & \textbf{Method} & \textbf{Train} & \textbf{Test} & \textbf{Evals} & \textbf{Tokens (M)} \\
\midrule
\multirow{6}{*}{CVRP-DR} 
 & Standard & $10.791$ & $10.558$ & -- & -- \\
 & MOTIF & $9.156 \pm 0.164$ & $9.204 \pm 0.114$ & $500$ & $317.1$ \\
 & E2OC & $9.920 \pm 0.078$ & $9.837 \pm 0.098$ & $500$ & $470.0$ \\
 & Synergy & $\underline{8.928 \pm 0.075}$ & $\underline{9.112 \pm 0.043}$ & $500$ & $289.6$ \\
 & Rotating-EoH & $8.998 \pm 0.067$ & $9.178 \pm 0.166$ & $500$ & $337.9$ \\
 & \systemname{} & \hpC{$8.860 \pm 0.037$} & \hpC{$8.999 \pm 0.036$} & \hpA{$271$} & \hpA{$139.2$} \\
\midrule
\multirow{6}{*}{FJSP 4-Ops} 
 & Expert & $712.5 \pm 5.0$ & $2137.9 \pm 12.0$ & -- & -- \\
 & MOTIF & $648.07 \pm 3.43$ & $\underline{1952.00 \pm 23.41}$ & $500$ & $584.7$ \\
 & E2OC & $663.47 \pm 7.62$ & $2019.20 \pm 44.53$ & $500$ & $699.3$ \\
 & Synergy & $\underline{646.10 \pm 2.95}$ & $1965.20 \pm 14.81$ & $500$ & $460.7$ \\
 & Rotating-EoH & $\mathbf{645.10 \pm 3.37}$ & $1954.87 \pm 26.14$ & $500$ & $518.5$ \\
 & \systemname{} & \hn{$646.4 \pm 3.3$} & \hpC{$1927.5 \pm 38.5$} & \hpA{$275$} & \hpA{$118.1$} \\
\bottomrule
\end{tabular}
\end{table}

\subsection{Main Results}
\label{sec:exp-main}

To test whether a single agentic engine can match or surpass paradigm-specialized baselines, we evaluate \systemname{} on all six tasks, changing only the activated design skill.

\paragraph{Single-heuristic design.}
On TSP and CVRP \systemname{} attains the best training and test quality, ahead of the strongest LLM-AHD baseline by $4$ to $7\%$ and of the nearest-neighbour heuristic by $12.3\%$ on TSP. The margins are not bought with budget, as the baselines run to the $500$-evaluation cap and remain behind, while \systemname{} finds its best incumbent by about $35$ evaluations, a thirteenth of the allowance.

\begin{figure}[t]
\centering
\begin{subfigure}[t]{0.49\linewidth}
\centering
\includegraphics[width=\linewidth]{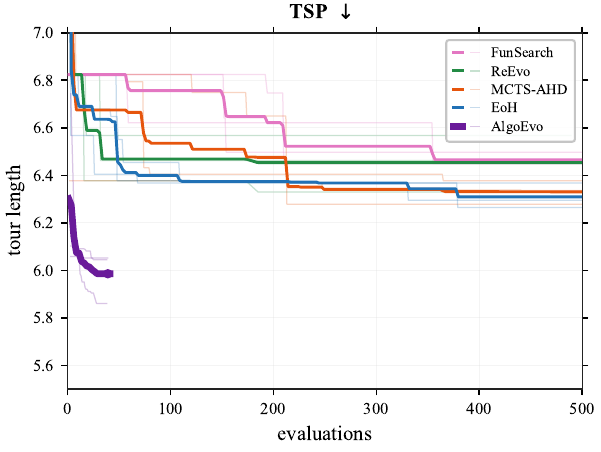}
\caption{TSP constructive heuristic design.}
\label{fig:budget-a}
\end{subfigure}
\hfill
\begin{subfigure}[t]{0.49\linewidth}
\centering
\includegraphics[width=\linewidth]{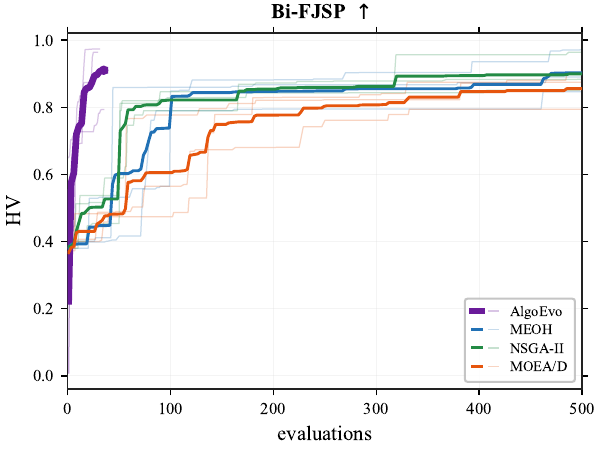}
\caption{Bi-FJSP multi-objective heuristic design.}
\label{fig:budget-b}
\end{subfigure}
\caption{Best-so-far quality as a function of the evaluation budget on TSP and Bi-FJSP. Thin lines are individual seeds; bold lines are means. \systemname{} reaches its final quality within the first tens of evaluations and then stays flat, whereas the baselines consume their full budget.}
\label{fig:budget}
\end{figure}

\paragraph{Multi-objective design.}
\systemname{} attains, on training instances, the best HV on both tasks and the best IGD on Bi-TSP, while trailing MEoH on front proximity for Bi-FJSP. The archives reveal that our front reaches low-objective corners missed by the baselines, whereas the MOEA/D front collapses to a point (Figure~\ref{fig:apppareto}).

\paragraph{Multi-component design.}
Here the components are coupled and no single unit suffices. \systemname{} obtains the best result of any method on both splits of CVRP-DR and the best held-out result on FJSP four-operator co-design, where it also improves on the expert operators. Unlike the four tasks above, it draws on most of the allowance rather than stopping early, so the margins come at or below the budget the baselines spend, and they are widest on the held-out sets, which is where a design has to transfer rather than to fit.

\subsection{Analysis and Discussion}
\label{sec:exp-analysis}

\paragraph{Experience verification.}
\setlength{\intextsep}{\wraptablesep}
\begin{wraptable}{r}{0.5\textwidth}
\centering
\footnotesize
\setlength{\tabcolsep}{4pt}
\caption{Effect of accumulated experience on TSP. \textbf{cold} starts from the default heuristic; \textbf{warm} from the best algorithm of a previous run.}
\label{tab:warm_start}
\begin{tabular}{@{}lcccc@{}}
\toprule
\multirow{2}{*}{\textbf{Seed}} & \multicolumn{2}{c}{\cellcolor{grpA}\textbf{Cold (no exp.)}} & \multicolumn{2}{c}{\cellcolor{grpC}\textbf{Warm (with exp.)}} \\
\cmidrule(lr){2-3} \cmidrule(lr){4-5}
& \textbf{Tour} & \textbf{Evals} & \textbf{Tour} & \textbf{Evals} \\
\midrule
2025 & $5.006$ & $45$ & $\mathbf{4.819}$ & $52$ \\
2026 & $4.879$ & $48$ & $\mathbf{4.810}$ & $28$ \\
2027 & $4.997$ & $51$ & $\mathbf{4.794}$ & $41$ \\
\midrule
Mean & $4.961$ & $48$ & $\mathbf{4.808}$ & $40$ \\
\bottomrule
\end{tabular}
\end{wraptable}
Within a task, the experience bank grows as the agent evaluates candidates, reusing earlier design decisions rather than exploring from scratch, so cost decreases as cards accumulate. The agent reaches its final quality well before the budget is exhausted. Experience also carries over between successive runs of the same task, where Table~\ref{tab:warm_start} shows that the warm-started agent consistently improves over the cold start while reducing the variance, indicating that transferring executable code is more effective than injecting abstract strategy descriptions.

\paragraph{Multi-round progressive accumulation}
\begin{wraptable}{r}{0.5\textwidth}
\centering
\footnotesize
\setlength{\tabcolsep}{4pt}
\caption{Performance across successive optimization rounds on TSP and Bi-objective FJSP.}
\label{tab:cumulative}
\begin{tabular}{@{}lcccc@{}}
\toprule
& \multicolumn{2}{c}{\cellcolor{grpA}\textbf{TSP constructive}} & \multicolumn{2}{c}{\cellcolor{grpB}\textbf{Bi-objective FJSP}} \\
\cmidrule(lr){2-3} \cmidrule(lr){4-5}
\textbf{Round} & \textbf{Tour} $\downarrow$ & $\boldsymbol{\Delta}$ & \textbf{HV} $\uparrow$ & $\boldsymbol{\Delta}$ \\
\midrule
0 & $4.8928$ & --- & $0.1549$ & --- \\
1 & $4.8011$ & $-1.88\%$ & $0.1688$ & $+8.97\%$ \\
2 & $4.8011$ & $-1.88\%$ & $0.1832$ & $+18.27\%$ \\
3 & $4.8003$ & $-1.89\%$ & $0.1840$ & $+18.79\%$ \\
4 & $\mathbf{4.7995}$ & $\mathbf{-1.91\%}$ & $\mathbf{0.1862}$ & $\mathbf{+20.21\%}$ \\
\bottomrule
\end{tabular}
\end{wraptable}
Table~\ref{tab:cumulative} isolates the contribution of the experience bank with identical budgets. Round $0$ initializes from a default heuristic with an empty bank, and each subsequent round resumes from the preceding best code, with $\boldsymbol{\Delta}$ measured against Round $0$. Both problems exhibit steady, monotonic improvement, advancing from $4.8928$ to $4.7995$ in tour length and $0.1549$ to $0.1862$ in HV. Because the evaluation budget remains fixed, this progressive enhancement confirms that accumulated experience actively guides search toward superior regions. Furthermore, promoting validated patterns into the design skill hub reinforces search stability (Algorithm~\ref{alg:skill-evolution}), and ablating the hub degrades performance (Appendix~\ref{app:ablation}).

\paragraph{Cross-task transfer.}
The experience bank facilitates knowledge transfer across problem settings. As shown in Table~\ref{tab:transfer}, transferring distilled patterns from single-objective CVRP to multi-component CVRP-DR allows \systemname{} to reuse edge-scoring rules within the constructor. This cross-task transfer enhances solution quality and lowers the mean cost from $10.314$ to $9.816$ (surpassing the cold-start baseline of $10.791$), demonstrating that the hierarchical experience bank enables rapid adaptation and superior performance on complex downstream tasks.

\setlength{\intextsep}{12pt plus 2pt minus 2pt}

\begin{table}[h]
\centering
\footnotesize
\setlength{\tabcolsep}{4pt}
\caption{Cross-problem experience transfer. A bank of validated design experience from the source task is made available to the target task, against a cold start that differs only in the absence of that experience; the better arm is \textbf{bold}.}
\label{tab:transfer}
\begin{tabular}{@{}llcccc@{}}
\toprule
\textbf{Source $\rightarrow$ target} & \textbf{Metric} & \textbf{Baseline} & \textbf{Cold start} & \textbf{With experience} & \textbf{Change}  \\
\midrule
CVRP $\rightarrow$ CVRP-DR & cost $\downarrow$ & $10.791$ & $10.314 \pm 0.345$ & $\mathbf{9.816 \pm 0.642}$ & $-4.8\%$ \\
TSP $\rightarrow$ Bi-TSP & HV $\uparrow$ & $152.2$ & $336.5 \pm 130.6$ & $\mathbf{352.2 \pm 141.6}$ & $+4.7\%$ \\
\bottomrule
\end{tabular}
\end{table}

\paragraph{Sample efficiency.}
\systemname{} reaches final quality within tens of evaluations without further improvement, whereas baselines keep progressing toward full budget without catching up (Figure~\ref{fig:budget}). That curvature is not an artefact of early stopping, since the agent keeps proposing changes until the harness limit is reached and few proposals improve on the incumbent; the flat tail therefore reflects the absence of further improvement rather than a deliberate stop.

\begin{figure}[h]
\centering
\includegraphics[width=\linewidth]{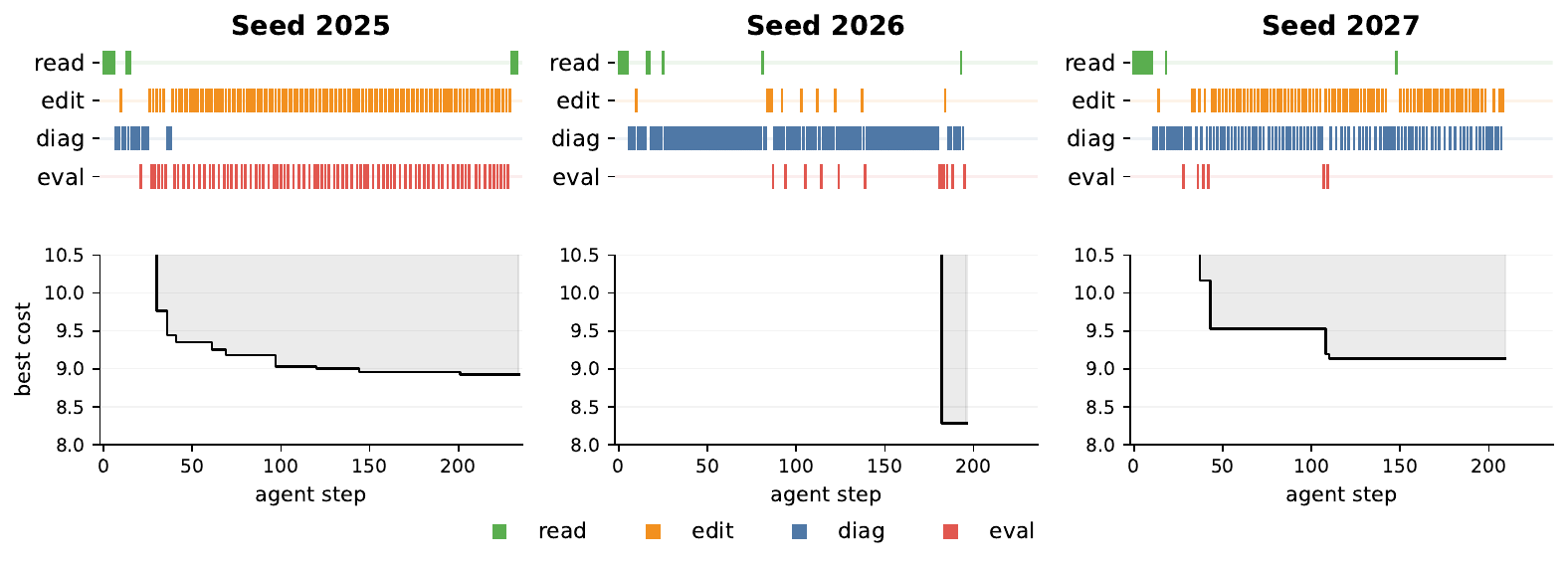}
\caption{Agent behavior trajectories on CVRP-DR across three seeds. Each row is an action type (evaluation, diagnose, edit, read); the black curve shows the best cost achieved so far.}
\label{fig:trajectories}
\end{figure}

\paragraph{Action behavior.}
We instrument the framework to log every tool invocation and analyze the resulting trajectories on CVRP-DR across three seeds. Each call is classified as read, edit, diagnose, or evaluation, and Figure~\ref{fig:trajectories} summarizes their evolution over time. The three seeds exhibit different search styles, one edit-driven, another diagnosis-driven, and the third mixing both. The diagnosis-driven seed reaches the best cost, consistent with the diagnosis-guided design principle, in which investing in structural attribution before editing avoids wasted modifications.

\subsection{Method Skills for Automated Algorithm Design}
\label{sec:methodskill}

The design skill abstraction is not limited to knowledge accumulated by \systemname{} itself; it can also carry methodological knowledge distilled from existing algorithm-design methods. We build method skills from six established approaches spanning single- and multi-objective design, and supply them to \systemname{} without importing their native search procedures. The skill-based variants match or exceed the native frameworks on solution quality while requiring substantially fewer evaluations, showing that methodological knowledge can be transferred and reused independently of the framework that produced it. The full protocol and results are given in Appendix~\ref{app:methodskill}.

\section{Conclusion}
\label{sec:conclusion}

In this paper, we introduced \systemname{}, a unified agentic framework that transforms automated algorithm discovery from rigid, expert-predefined pipelines into an adaptive, knowledge-accumulating search process. By decoupling paradigm-specific knowledge through the design skill hub and organizing discovery trajectories via the hierarchical experience bank, \systemname{} enables a single engine to seamlessly govern single-heuristic, multi-objective, and multi-component tasks. Extensive empirical evaluations across six representative combinatorial optimization benchmarks demonstrate that our state-aware search control achieves state-of-the-art solution quality while consuming as little as 7\% of the standard evaluation budget. Furthermore, our findings highlight the efficacy of intra-task experience accumulation and cross-task knowledge transfer in bootstrapping complex algorithmic co-design. This work establishes a principled and scalable foundation for self-evolving automated algorithm discovery, opening new avenues for intelligent scientific optimization.

% \subsection*{AI use statement}
% Generative AI tools were used to aid and polish the writing of this paper. All AI-generated content was reviewed and verified by the authors, who take responsibility for the final content of this work.

% \subsection*{Reproducibility statement}
% We document the parameter settings and the concrete implementation of the framework in the appendix, including the pseudocode of the discovery loop and the benchmark procedures. The code is provided to the reviewers as supplementary material and will be released publicly upon acceptance.

% \subsection*{Ethics statement}
% This work studies automated algorithm design on standard combinatorial optimization benchmarks. It does not involve human subjects, personal or sensitive data, or deployment in a safety-critical setting, and we are not aware of ethical concerns that it raises.

\bibliography{iclr2027_conference}
\bibliographystyle{iclr2027_conference}

% ---------------------------------------------------------------------------
% Drafting aid: appendix-only contents list on a page of its own. The main text
% is suppressed by the tocdepth marker set after \maketitle, and the appendix
% entries are re-enabled by the marker set after \appendix. Remove this block
% and both markers before submission.
% ---------------------------------------------------------------------------
{
\renewcommand{\contentsname}{Appendix Contents}
\setcounter{tocdepth}{2}
\clearpage
\begin{center}
{\Large\bfseries\papertitle\par}
\end{center}
\vspace{0.5em}
\tableofcontents
\clearpage
}

% ============================================================================
% APPENDIX: Related Work is migrated from V0; Sections B to D are still an
% outline to be filled.
% Aligned with the main text of this manuscript (Sections 2 to 4).
% Planned sources: manuscriptV0.tex appendices B to D, and the released
% implementation under algoevo/.
% Each appendix section must answer something the main text raises but does not
% settle. Mechanism already stated in Sections 3 is NOT repeated here.
% NOTE: tcolorbox is now loaded in the preamble, so the unit code boxes of C.4
% render; no further package is needed for verbatim blocks.
% ============================================================================
\appendix

\addtocontents{toc}{\protect\setcounter{tocdepth}{2}}

% ============================================================
% B. Related Work
% ============================================================
\section{Related Work}
\label{app:related}

\systemname{} lies at the intersection of automated algorithm design, agentic software engineering, and knowledge-guided search. We therefore review related work from both the algorithm-design and general agent perspectives. The fundamental distinction is not merely whether an approach utilizes language models, tools, memory, or skills, but what is being searched, how the search state is defined, and how empirical feedback governs subsequent actions.

\paragraph{LLM-based Automated Algorithm and Heuristic Design.}
LLMs have recently enabled a powerful paradigm for automated algorithm and heuristic design \citep{llm4ad2026survey}. FunSearch combines LLM-based program generation with evolutionary search, demonstrating the capacity of language models to discover programs with improved performance \citep{funsearch2024}. EoH formulates heuristic design as an iterative process of generating heuristic ideas and implementations and progressively improving them through evaluation \citep{eoh2024}. ReEvo further introduces evolutionary refinement of LLM-generated heuristics, enabling iterative improvement of previously discovered designs \citep{reevo2024}. MCTS-AHD employs Monte Carlo tree search to structure the exploration of algorithm-design candidates \citep{mctsahd2025}. In multi-objective settings, MEoH extends LLM-based heuristic design to multi-objective evolutionary search \citep{meoh2025}. Additional recent approaches include EvoPrompt, which connects LLMs with evolutionary algorithms for prompt optimization \citep{guo2024connecting}, LLaMEA, which uses LLMs to automatically generate metaheuristics \citep{llamea2025}, and HSEvo, which combines harmony search with genetic algorithms via LLMs \citep{hsevo2025}. EoH-S further extends EoH to heuristic set design \citep{eohs2026}. These studies establish that LLMs can serve as effective generators and refiners of algorithmic ideas and executable implementations. However, their underlying search procedures and multi-component coordination policies remain bound to fixed, predefined protocols. Traditional automated configuration (e.g., irace \citep{lopez2016irace}) and recent multi-component frameworks like MOTIF \citep{motif2025} and E2OC \citep{e2oc2026} demonstrate that search strategies and cross-component interactions are critical, yet they typically instantiate coordination via static, algorithm-specific rules rather than adaptive control.

\paragraph{Agentic Coding and Program Synthesis.}
Recent advances in LLM-based coding agents (e.g., SWE-agent \citep{yang2024swe} and MetaGPT \citep{metagpt2024}) demonstrate autonomous code generation, repository exploration, and iterative debugging, while performance-driven evolutionary coding agents such as AlphaEvolve \citep{alphaevolve2025} optimize programs against empirical objectives \citep{agent2024survey}. Both, however, operate over concrete programs rather than algorithm-design knowledge: general-purpose software agents are driven by functional correctness, syntactic validity, and test compliance against static specifications, while evolutionary coding agents search over program variants without an explicit model of the algorithm being designed. In algorithm discovery, by contrast, a syntactically flawless program frequently exhibits severe optimization bottlenecks, poor search trajectories, or weak generalization across instances. Bridging across complex algorithmic paradigms requires navigating intricate search semantics and multi-component interactions that cannot be solved simply by employing a strong general-purpose coding agent. \systemname{} overcomes this limitation by treating code merely as the implementation medium, whereas empirical optimization feedback across diverse problem instances acts as the primary state signal guiding structural redesign. The agentic interface is thus elevated into an optimization-driven design search, rather than functioning as a standard coding assistant.

\paragraph{Skill-based and Experience-driven Agents.}
A growing body of research equips LLM agents with reusable skills, external memory, and behavioral evolution mechanisms \citep{li2026dynamic}, ranging from single-agent workflow memory systems such as AWM \citep{awm2024}, G\"odel Agent \citep{godelagent2025}, and Trace2Skill \citep{trace2skill2026}, to multi-objective skill bundle optimization frameworks such as SkillMOO \citep{gong2026skillmoo}. VLM agents have also demonstrated the ability to generate their own memories by distilling experience into embodied programs of thought \citep{sarch2024vlm}, while procedural memory frameworks enable experience-driven agent evolution through dynamic refinement \citep{cao2026remember}. Memory evolution surveys further demonstrate the progression from static storage to experience-driven agent improvement \citep{luo2026storage}. For skill creation and transfer, Trace2Skill distills trajectory-local lessons into transferable agent skills \citep{trace2skill2026}, while Ctx2Skill explores how language models learn context-specific skills through self-play \citep{ctx2skill2026}. However, recent empirical studies reveal critical limitations of generic skill representations: skill libraries can degrade agent performance by up to 21\% as they expand due to skill shadowing effects \citep{song2026more}, over 89\% of real-world skills exhibit reusability defects \citep{zhang2026keeps}, and skills can cause regressions through description osmosis, grounding displacement, and verification displacement \citep{tank2026regression}. Moreover, indiscriminate memory addition can propagate errors and contaminate agent learning \citep{xiong2026memory}, while agentic memory systems that enable dynamic organization show superior performance over static approaches \citep{xu2026mem}. These findings confirm that such generic structures fail to capture the specialized semantics required for algorithm optimization. Algorithm discovery is uniquely challenging because design choices exhibit deep structural coupling \citep{chen2026diversity}, and historical performance trends are non-linear. \systemname{} addresses this by transforming these mechanisms into rigorous, domain-specific algorithm-design knowledge. A Design Skill encodes execution mechanisms, applicable conditions, design principles, and characteristic failure or convergence patterns. Similarly, the Experience Bank records validated design decisions alongside their empirical optimization effects, establishing a continuous knowledge cycle connecting experience acquisition, knowledge abstraction, and state-aware design search. This specialized structure ensures that knowledge governs not only text retrieval, but also the dynamic selection and coordination of subsequent algorithmic components.

\paragraph{Positioning of AlgoEvo.}
The three lines of work above leave three distinct gaps, each addressed by one component of \systemname{}. First, LLM-based algorithm discovery searches candidate programs under fixed protocols, so the design process itself is never searched; \systemname{} answers this with its \emph{agentic search loop}, in which the agent chooses among inspection, diagnosis, modification, and evaluation according to the current optimization state rather than following a prescribed cycle. Second, general-purpose coding agents and evolutionary program searchers act on concrete programs without a representation of the algorithm under design, and cross-component coordination is either absent or encoded in static, algorithm-specific rules; \systemname{} answers this with its \emph{design skill hub}, whose paradigm-specific contracts let a single engine span single-heuristic, multi-objective, and multi-component tasks. Third, generic skill and memory structures, as the empirical studies above indicate, do not capture the semantics of algorithm design and can even degrade as they grow; \systemname{} answers this with its \emph{hierarchical experience bank}, which records design decisions as situation-conditioned cards, organizes them into a task-level tree, and consolidates validated patterns into transferable skills. Together, these components recast algorithm discovery from a fixed candidate-generation procedure into an optimization-driven, state-aware search over the design process.

% ============================================================
% C. Additional Method Details
% ============================================================

\section{Method Details}
\label{app:method}

This section provides a concrete, implementation-level description of the algorithm discovery process introduced in Section~\ref{sec:algoevo}. Rather than executing a predefined script of optimization steps, \systemname{} treats algorithm design as an interactive agentic search process. The framework coordinates agent reasoning, diagnostic feedback, methodological guidance, and hierarchical experience accumulation to explore complex algorithmic spaces under a strict evaluation budget. We detail the task specification and initialization, the discovery state, actions, and diagnosis, the design skill hub, the experience bank, and the concrete artifacts and prompts used in the implementation.

For reference, Algorithm~\ref{alg:algoevo-full} restates the complete discovery loop, expanding the condensed version in the main text (Algorithm~\ref{alg:algoevo}) into its constituent steps: situation categorization, experience retrieval via UCB, experience-card construction, reward backpropagation, and cross-task skill evolution.

\begin{algorithm}[!ht]
\caption{Agentic Algorithm Discovery (full)}
\label{alg:algoevo-full}
\begin{algorithmic}[1]
\Require Design task $\mathcal{T}$, design skill hub $\mathcal{S}_{\mathrm{hub}}$, experience bank $\mathcal{E}$, budget $B$, threshold $\mu$
\Ensure Optimized algorithm $\mathbf{\Pi}^{*}$
\State $\sigma \leftarrow\mathrm{\textbf{Activate}}(\mathcal{T},\mathcal{S}_{\mathrm{hub}})$ \Comment{Skill activation (Algo. ~\ref{alg:init})}
\State $\epsilon \leftarrow\mathrm{\textbf{Match}}(\mathcal{T},\sigma,\mathcal{E})$ \Comment{Experience tree matching (Algorithm~\ref{alg:init})}
\State $t \leftarrow 0$ \Comment{Step counter}
\State Initialize algorithm $\mathbf{\Pi} \leftarrow \mathbf{\Pi}_0$, working context $\mathcal{C}_t$, and compute $F_d(\mathbf{x}\mid\mathbf{\Pi}_0)$
\Repeat
    \State \textit{// 1. Observe: detect current situation from global discovery state}
    \State Update global discovery state $s_t \leftarrow (\mathbf{\Pi}, \mathcal{C}_t, \epsilon, F_d(\mathbf{x}\mid \mathbf{\Pi}))$
    \State $q_t \leftarrow \mathrm{\textbf{Categorize}}(s_t)$ \Comment{Extract conditions (Algorithm~\ref{alg:app-situation})}
    \State \textit{// 2. Action: The agent selects an action, such as inspect, retrieve, diagnose, edit, or evaluate.}
    \State $a_t \leftarrow \mathrm{\textbf{Agent}}(\sigma, q_t, \epsilon, \mathcal{C}_t)$
    \State \textit{// 3. Update: process outcome}
    \If{$a_t = \text{evaluate}$}
        \State Obtain the latest algorithm candidate $\mathbf{\Pi}'$ and compute reward $r_t$ \Comment{Eq.~\ref{eq:reward}}
        \State $\mathbf{\Pi} \leftarrow \mathbf{\Pi}'$ \Comment{Update current incumbent algorithm}
        \State Create card $c_t \leftarrow \mathrm{\textbf{Summarize}}(\mathcal{T}, \sigma, a_t, F_d(\mathbf{x}\mid \mathbf{\Pi}))$
        \State Add $c_t$ to $\epsilon$ and backpropagate $r_t$ \Comment{Card $\leftrightarrow$ node mapping}
        \State Update working context $\mathcal{C}_t$
    \ElsIf{$a_t = \text{diagnose}$ or $a_t = \text{retrieve}$}
        \State Select experience $e$ via UCB: $e \leftarrow \arg\max_{e' \in \epsilon} \mathrm{UCB}(e' \mid q_t)$ \Comment{Eq.~\ref{eq:ucb}}
        \State Agent reads $e$ and updates working context $\mathcal{C}_t$
    \EndIf
    \State $t \leftarrow t + 1$
\Until{budget $B$ exhausted or $a_t = \text{terminate}$}
\State $\mathbf{\Pi}^{*} \leftarrow \arg\min_{\mathbf{\Pi}} F_d(\mathbf{x}\mid \mathbf{\Pi})$ \Comment{Select best incumbent over evaluated candidates}
\State $\mathrm{\textbf{Evolve}}(\sigma, \mathcal{E}, \mathcal{S}_{\mathrm{hub}}, \mu)$ \Comment{Cross-task consolidation (Algorithm~\ref{alg:skill-evolution})}
\State \Return $\mathbf{\Pi}^{*}$
\end{algorithmic}
\end{algorithm}

\subsection{Task Specification and Initialization}
\label{app:method-task}

A design task is fully specified at initialization by three artifacts: the initial algorithm code, the set of evolvable units, and the evaluation function $F_d$, together with a budget $B$ equal to the maximum number of evaluations allowed for the task. These are fixed for the entire run and shared by all agent actions, so the search mechanism itself is paradigm-agnostic. A \emph{multi-component} task simply declares an algorithm $\mathbf{\Pi}=(\pi_1,\ldots,\pi_K)$ whose strategies decompose into coupled sub-components; the agent edits those sub-components directly, and the diagnostic engine measures their coupling. A \emph{multi-objective} task differs only in the evaluation function, which returns a vector of objectives rather than a scalar; the reward of Eq.~\ref{eq:reward} is then computed on the aggregated objective value, which for these tasks is the negative hypervolume of the current Pareto front under the task's reference point. In both cases, supporting the paradigm requires only a different task specification; the loop itself is unchanged.

\paragraph{Initialization.}
\label{app:method-init}

At task initialization, the system generates a system prompt that states the method and the task (Section~\ref{app:method-prompt}). Guided by this prompt, the agent activates the design skill whose declared conditions best match the task, and matches an experience tree semantically indexed by (skill, problem, evaluator). If no matching tree exists, a new one is created. Algorithm~\ref{alg:init} formalizes this process.

\begin{algorithm}[!ht]
\caption{Task Initialization}
\label{alg:init}
\begin{algorithmic}[1]
\Require Design task $\mathcal{T}$, skill hub $\mathcal{S}_{\mathrm{hub}}$, experience bank $\mathcal{E}$
\Ensure Algorithm design skill $\sigma$, experience tree $\epsilon$
\State Generate system prompt $P$ based on problem description, objectives, and design units
\State Inject $P$ into agent
\State Agent analyzes $P$ and activates appropriate skill $\sigma$ from $\mathcal{S}_{\mathrm{hub}}$ \Comment{Activate}
\State Agent matches an experience tree semantically indexed by $(\mathcal{T}.\text{problem}, \mathcal{T}.\text{evaluator}, \sigma)$ in $\mathcal{E}$ \Comment{Match}
\If{a matching tree is found}
    \State Retrieve existing tree $\epsilon$
\Else
    \State Create new experience tree $\epsilon$ and store it
\EndIf
\State \Return $\sigma, \epsilon$
\end{algorithmic}
\end{algorithm}

\subsection{Discovery State, Actions, and Diagnosis}
\label{app:method-state}

At each step of the discovery trajectory, the agent maintains access to a comprehensive state space comprising the current algorithm implementation, working context, historical decisions, diagnostic feedback, and performance metrics. The working context logs all operations executed within the current trajectory, including inspected code segments, attempted modifications, evaluation outcomes, and analytical conclusions. This distinction prevents the agent from conflating a fresh optimization problem with the continuation of an unsuccessful branch.

Discovery actions are formulated at the level of high-level design decisions. Table~\ref{tab:actions-app} itemizes them and their resource consumption. A critical architectural distinction separates evaluation from analytical actions, in that only evaluation yields a candidate algorithm and consumes the evaluation budget, whereas diagnostic and relational analyses extract structural evidence without producing candidates and without charging that budget. Analytical actions are nevertheless bounded by the overall trajectory limit, so the agent must balance empirical testing against analytical reasoning.

\begin{table}[!ht]
\centering
\footnotesize
\setlength{\tabcolsep}{4pt}
\caption{Discovery actions available to the agent, their state impacts, and resource consumption.}
\label{tab:actions-app}
\begin{tabular}{@{}l>{\raggedright\arraybackslash}p{0.40\linewidth}cc@{}}
\toprule
\textbf{Action} & \textbf{Effect on the discovery state} & \cellcolor{grpA}\textbf{Candidate} & \cellcolor{grpC}\textbf{Charge} \\
\midrule
Inspect & Reads source files or task execution artifacts & No & --- \\
Retrieve & Queries the experience bank and reasoning chains & No & --- \\
Diagnose & Attributes performance to units and identifies failure modes & No & --- \\
Analyse interaction & Measures coupling strength between two units & No & --- \\
Edit & Modifies the source code of a target design unit & No & --- \\
Evaluate & Executes and scores the current on-disk algorithm & \cellcolor{grpA}Yes & \cellcolor{grpC}Sample, eval. \\
Commit & Persists validated high-performing versions & \cellcolor{grpA}Yes & \cellcolor{grpC}Eval. \\
Terminate & Concludes the active task trajectory & No & --- \\
\bottomrule
\end{tabular}
\end{table}

The resulting trajectory forms a heterogeneous sequence of analytical, exploratory, and evaluative steps. In multi-component tasks, this enables the agent to isolate interacting units, measure their coupling, and perform coordinated edits.

Because the trajectory mixes evaluative and analytical steps, the reward is defined only for steps that update the incumbent algorithm, relative to the candidate from which the update was derived (its parent card when the agent branches from an earlier node), and is taken as zero otherwise. The incumbent tracks the latest evaluated candidate, while the algorithm returned at the end is the best incumbent over the whole trajectory, so that a temporarily inferior candidate does not overwrite the final answer.

\paragraph{Diagnosis and situation detection.}
\label{app:method-diag}

Relying solely on scalar performance scores provides insufficient guidance for complex search spaces. While a low evaluation score indicates that an algorithm is suboptimal, it fails to specify which component caused the failure or whether multiple units interact adversely. To overcome this, \systemname{} incorporates situation-aware retrieval to match historical experiences with the current search context.

At each step, the diagnostic engine supplies two runtime signals, a per-unit attribution $A_u$ quantifying the contribution of design unit $u$ to the current performance gap and a scalar interaction measurement $\mathrm{int}$ capturing the coupling between units. Together with the submission history and the Pareto archive, these signals determine the active situation set $q_t\subseteq\mathcal{Q}$, where $\mathcal{Q}$ collects the situation categories (stagnation, bottleneck, coupling, sparse front). Algorithm~\ref{alg:app-situation} formalizes this process, in which Part~1 categorizes $q_t$ from the discovery state $s_t$ and Part~2 selects the experience card $e$ that best matches it. Thresholds $\tau_s$, $\tau_c$, and $\tau_b$ are listed in Table~\ref{tab:situations}.

The engine computes the two signals by counterfactual evaluation on the training set. The per-unit attribution replaces the body of unit $u$ with a neutral no-op implementation and records the resulting change in the evaluation score,
\begin{equation}
A_u = F_d(\mathbf{x}\mid\mathbf{\Pi}_{u\leftarrow\mathrm{noop}}) - F_d(\mathbf{x}\mid\mathbf{\Pi}),
\label{eq:attribution}
\end{equation}
so that the unit whose neutralization degrades performance most is the dominant contributor to the current gap. The interaction measurement neutralizes a pair of units jointly and applies inclusion-exclusion,
\begin{equation}
\mathrm{int}(u_a,u_b) = A_{u_a\wedge u_b} - A_{u_a} - A_{u_b},
\label{eq:interaction}
\end{equation}
where $A_{u_a\wedge u_b}$ is the attribution obtained by neutralizing both units at once; a large magnitude indicates strong coupling, so the two units are better redesigned jointly. Both probes are analysis operations rather than candidate submissions, producing no candidate and consuming no evaluation budget, although the trajectory remains bounded by the step limit. The remaining conditions are read directly from the search history, namely stagnation after $\tau_s$ consecutive non-improving submissions and a sparse front when the normalized archive has at least two points whose largest nearest-neighbour gap exceeds three times the smallest.

\begin{algorithm}[!ht]
\caption{Situation categorization and experience card selection}
\label{alg:app-situation}
\begin{algorithmic}[1]
\Require Discovery state $s_t$, experience tree $\epsilon$
\Ensure Active situations $q_t$, selected experience card $e$
\State \textit{// Part 1: Categorize the current situation from $s_t$}
\State $q_t \leftarrow \emptyset$
\If{last $\tau_s$ submissions show no improvement}
    \State $q_t \leftarrow q_t \cup \{\text{stagnation}\}$
\EndIf
\If{$|\mathrm{int}| > \tau_c$}
    \State $q_t \leftarrow q_t \cup \{\text{coupling}\}$
\EndIf
\If{$\max_u A_u / \sum_u A_u > \tau_b$}
    \State $q_t \leftarrow q_t \cup \{\text{bottleneck}\}$
\EndIf
\If{front has wide gaps}
    \State $q_t \leftarrow q_t \cup \{\text{sparse front}\}$
\EndIf
\State \textit{// Part 2: Select the experience card}
\State $e \leftarrow \arg\max_{e'\in\epsilon}\mathrm{UCB}(e'\mid q_t)$ \Comment{Select the highest-scoring card by Eq.~\ref{eq:ucb}}
\State \Return $q_t, e$
\end{algorithmic}
\end{algorithm}

Algorithm~\ref{alg:app-situation} instantiates the \textbf{Categorize} step of Algorithm~\ref{alg:algoevo}: it detects the active situations $q_t$ from the discovery state $s_t$ (Part 1) and then selects the experience card $e$ that best matches them (Part 2). The highest-scoring card by Eq.~\ref{eq:ucb} guides the next design decision.

\begin{table}[!ht]
\centering
\footnotesize
\setlength{\tabcolsep}{4pt}
\caption{Discovery situations and their activation conditions.}
\label{tab:situations}
\begin{tabular}{@{}ll@{}}
\toprule
\textbf{Situation} & \textbf{Activated when} \\
\midrule
\textbf{Stagnation} & The last $\tau_s = 4$ submissions yield no improvement \\
\textbf{Sparse front} & The largest nearest-neighbour gap in the normalized archive exceeds $3\times$ the smallest \\
\textbf{Bottleneck} & A single unit $u$ accounts for $> \tau_b = 0.5$ of the attribution mass \\
\textbf{Coupling} & The interaction measurement satisfies $|\mathrm{int}| > \tau_c = 0.05$ \\
\bottomrule
\end{tabular}
\end{table}

This diagnostic mechanism decouples behavioral understanding from raw objective scoring, ensuring that modifications target verified structural weaknesses.

\subsection{Design Skill Hub}
\label{app:method-skill}

Different algorithmic paradigms impose unique structural constraints and interface requirements. A constructive heuristic, a multi-objective operator set, and a coupled destroy-and-repair pipeline require distinct design strategies. The design skill hub provides paradigm-specific methodological knowledge without altering the core agentic search loop.

\paragraph{Skill representation.}
Each design skill is stored as a Markdown file with YAML frontmatter (Listing~\ref{lst:skill-example}). The frontmatter declares the skill's metadata, applicable conditions, strategy roles, and operator taxonomy. The Markdown body provides detailed design guidance, failure modes, and convergence patterns.

\begin{figure}[!ht]
\centering
\begin{codebox}
\begin{lstlisting}[style=boxed]
---
name: eoh
version: 0.2.0
(*@\textbf{paradigm: population-based}@*)
applicable:
  problem_type: [single-objective, multi-objective]
  instance_set: [combinatorial, scheduling]
  solution_type: [permutation, assignment]
strategy_roles:
  pi_1:
    name: population_generator
    kind: generator
    file: generate.py
    signature: "def generate(problem, config) -> list"
  pi_2:
    name: solution_evaluator
    kind: evaluator
    file: evaluate.py
    signature: "def evaluate(problem, solution) -> float"
operators:
  perturbation: [random_reset, swap_mutation]
  hybridization: [uniform_crossover]
  repair: [feasibility_repair]
  improvement: [local_search]
---

(*@\textbf{\# EoH Design Skill}@*)

(*@\textbf{\#\# Design Principles}@*)
1. Maintain diversity through adaptive mutation rates
2. Balance exploration (perturbation) and exploitation
3. Use evaluation feedback to guide operator selection

(*@\textbf{\#\# Operator Taxonomy}@*)
- (*@\textbf{E1 (Perturbation)}@*): random changes to escape local optima
- (*@\textbf{E2 (Hybridization)}@*): combine features of two solutions
- (*@\textbf{M1 (Repair)}@*): restore feasibility after perturbation
- (*@\textbf{M2 (Improvement)}@*): local search to refine solutions

(*@\textbf{\#\# Failure Modes}@*)
- Over-reliance on perturbation leads to random walk
- Insufficient repair causes infeasible solutions
- Weak improvement operators slow convergence
\end{lstlisting}
\end{codebox}
\caption{Example skill representation for the EoH paradigm. The YAML frontmatter declares metadata, applicable conditions, and operator taxonomy; the Markdown body provides design guidance.}
\label{lst:skill-example}
\end{figure}

\paragraph{Skill matching.}
At task initialization, the agent selects the appropriate skill based on the task description (Algorithm~\ref{alg:init}). If the task config explicitly specifies a skill name, that skill is used directly. Otherwise, the agent analyzes the task requirements and selects the most suitable skill from the registry.

\paragraph{Skill evolution.}
Skills are not updated within a single task; evolution is triggered only across tasks. Algorithm~\ref{alg:skill-evolution} collects the experience trees owned by the current skill $\sigma$ and, once at least $\mu$ trees have accumulated ($\mu=2$ by default), has the agent summarize their most salient patterns and refine the skill with the summary. Because the refinement is grounded in accumulated cross-task evidence rather than a single trajectory, it sharpens the skill's methodological knowledge without overwriting its paradigm-level guidance.

\begin{algorithm}[!ht]
\caption{Cross-task skill evolution}
\label{alg:skill-evolution}
\begin{algorithmic}[1]
\Require Skill $\sigma$, experience bank $\mathcal{E}$, skill hub $\mathcal{S}_{\mathrm{hub}}$, threshold $\mu$
\Ensure Refined skill $\sigma$
\State Collect the experience trees in $\mathcal{E}$ owned by $\sigma$
\If{the number of collected trees is at least $\mu$}
    \State Agent summarizes the most salient patterns across the collected trees
    \State Refine $\sigma$ with the summary
\EndIf
\State \Return $\sigma$
\end{algorithmic}
\end{algorithm}

Extending the framework to new optimization paradigms requires defining corresponding design skills rather than restructuring the agent architecture. Furthermore, the hub incorporates distilled methodological insights from established optimization literature, equipping the agent with established design patterns and known failure modes.

\subsection{Hierarchical Experience Bank}
\label{app:method-bank}

While working contexts manage single-trajectory reasoning, the hierarchical experience bank preserves empirical knowledge across tasks. Evaluated candidates are distilled into structured experience cards recording the design context, applied modifications, empirical performance, and underlying rationale.

\paragraph{Experience card structure.}
Each experience card $e$ is a tuple $e=(\texttt{problem},\texttt{situation},\texttt{mode},\texttt{unit},\texttt{skill},\texttt{content},\texttt{evidence},\texttt{code},\texttt{metrics})$ that captures the full context of a design decision. The fields are defined in Table~\ref{tab:experience-card}. The \texttt{mode} field distinguishes validated strategies (\texttt{validated}), known failures (\texttt{avoid}), and post-task reflections (\texttt{reflection}). The \texttt{code} field optionally stores reference implementation that can be injected into the agent's context.

\begin{table}[!ht]
\centering
\footnotesize
\setlength{\tabcolsep}{4pt}
\caption{Experience card fields and their descriptions.}
\label{tab:experience-card}
\begin{tabular}{@{}ll>{\raggedright\arraybackslash}p{0.75\linewidth}@{}}
\toprule
\textbf{Field} & \textbf{Type} & \textbf{Description} \\
\midrule
\texttt{problem} & string & Problem name (e.g., ``TSP'', ``CVRP'') \\
\texttt{situation} & string & Detection context: initial, stagnation, bottleneck, coupling, sparse front, default \\
\texttt{mode} & string & Validated, avoid, or reflection \\
\texttt{unit} & string & Design unit this experience pertains to \\
\texttt{skill} & string & Owning paradigm skill name \\
\texttt{content} & string & Human-readable description of the design pattern \\
\texttt{evidence} & string & Empirical evidence (e.g., ``$\times 3$ improved, swing $\sim 0.15$'') \\
\texttt{code} & string & Optional reference implementation \\
\texttt{metrics} & dict & Optional performance metrics \\
\bottomrule
\end{tabular}
\end{table}

\paragraph{Situation-aware retrieval.}
Experience cards are selected by the situation-conditioned upper confidence bound of Eq.~\ref{eq:ucb} (Algorithm~\ref{alg:app-situation}). Each card maintains a mean reward and a visit count for the current situation, so that experiences validated under similar situations are preferred while cards with few observations retain an exploration incentive.

\paragraph{Experience tree.}
Experiences generated within a task are organized into a task-level experience tree, optimized via four Monte Carlo Tree Search stages:

\begin{enumerate}[leftmargin=*,itemsep=2pt,topsep=2pt]
\item \textit{Selection}: Balances exploitation of high-performing design branches with exploration of unvisited directions via the situation-conditioned upper confidence bound (Eq.~\ref{eq:ucb}).
\item \textit{Expansion}: Appends newly evaluated design decisions to the tree structure.
\item \textit{Evaluation}: Grounds experience nodes in empirical performance rather than language model heuristics.
\item \textit{Backpropagation}: Propagates performance outcomes up ancestral nodes, crediting foundational design choices that enabled downstream improvements.
\end{enumerate}

Querying the experience tree using active discovery situations ensures that retrieved historical insights match the current operational context.

\paragraph{Experience consolidation and promotion.}
\label{app:method-promotion}

To enable cumulative learning, the framework converts evaluation outcomes into reusable methodological knowledge. This section details the experience update and the reflection mechanism.

\paragraph{Experience update.}
The agent updates the experience bank only when it performs an evaluation, each of which drives it to summarize the outcome into an experience card, which is classified as validated or avoid and attached to the experience tree. Algorithm~\ref{alg:distill} formalizes this process.

\begin{algorithm}[!ht]
\caption{Experience card construction}
\label{alg:distill}
\begin{algorithmic}[1]
\Require Evaluation outcome of candidate $\mathbf{\Pi}$, active situation $q_t$, experience tree $\epsilon$
\Ensure Experience card $c_t$
\State Agent summarizes the evaluation into a card $c_t$, recording the situation $q_t$, the modification, the reward $r_t$, and the rationale
\State Set the mode of $c_t$ to validated or avoid
\State Attach $c_t$ to $\epsilon$
\State \Return $c_t$
\end{algorithmic}
\end{algorithm}

\paragraph{Post-task reflection.}
After the task completes, the agent generates a structured reflection summarizing the design process. Listing~\ref{lst:reflection} shows the reflection template, which captures effective process paths, situation-strategy mappings, and practices to avoid.

\begin{figure}[!ht]
\centering
\begin{codebox}
\begin{lstlisting}[style=boxed]
(*@\textbf{\#\#\# Effective Process Path}@*)
- step-by-step process that worked (numbered list, max 6 items)

(*@\textbf{\#\#\# Situation-Strategy Mapping}@*)
- for each situation encountered, which information/strategy helped
  (bullet list: "situation -> strategy")

(*@\textbf{\#\#\# Avoid List}@*)
- practices that failed or wasted budget (bullet list, max 5 items)
\end{lstlisting}
\end{codebox}
\caption{Reflection template used after each task. The reflection is stored as an experience card with mode=\texttt{reflection}.}
\label{lst:reflection}
\end{figure}

\paragraph{Cross-task promotion.}
A skill is refined only once it owns at least $\mu$ experience trees accumulated across tasks. This closes the self-evolving loop, allowing \systemname{} to accumulate transferable algorithmic intelligence over successive problem instances.

\subsection{Concrete Artifacts and Prompts}
\label{app:example}

We close the method details with concrete instances of the three knowledge artifacts the loop produces and consumes: a \emph{paradigm skill}, a \emph{method skill}, and an \emph{experience card}. A paradigm skill is keyed to the design setting (single-heuristic construction, multi-objective design, or multi-component co-design) and fixes how a task of that setting is designed (Section~\ref{sec:hub}); a method skill instead carries the methodology of an existing algorithm and is supplied without importing its native search loop (Section~\ref{sec:methodskill}); and an experience card is the unit of empirical evidence (Section~\ref{sec:expbank}).

\paragraph{Paradigm skill.}
A paradigm skill declares the roles of the designable strategies, their interfaces, the principles that guide valid modifications, and the evaluation conventions, together with the structural features that decide its activation. Listing~\ref{lst:skill-paradigm} shows the skill for the single-heuristic setting on TSP: its \texttt{paradigm} field names the setting, its \texttt{features} state that the task is single-objective with one function-typed unit in the \texttt{tsp} domain, and its body fixes the unit interface and the design knowledge.

\begin{figure}[!ht]
\centering
\begin{codebox}
\begin{lstlisting}[style=boxed]
---
name: tsp-constructive
version: 1.0.0
(*@\textbf{paradigm: single-heuristic}@*)
triggers: [tsp, travelling salesman, tour]
features:
  n_objectives: 1
  n_units: 1
  unit_kinds: [function]
  domain: [tsp]
---

(*@\textbf{\# TSP Constructive Skill}@*)

(*@\textbf{\#\# Applicability}@*)
Single unit, constructive; the evolvable unit is
  def select_next_node(current_node, destination_node,
                       unvisited_nodes, distance_matrix) -> int

(*@\textbf{\#\# Design Knowledge}@*)
- The nearest-neighbour rule is the reference point.
- Combining distance with a lookahead term improves on it.
- Caching and lookahead help at larger instance sizes.

(*@\textbf{\#\# Failure Modes}@*)
- A non-integer or out-of-range node crashes the evaluator.
- Pure distance scoring degenerates to nearest-neighbour.
- Over-complex scoring overfits the training instances.

(*@\textbf{\#\# Acceptance}@*)
- Valid signature; always returns an unvisited node.
\end{lstlisting}
\end{codebox}
\caption{A paradigm skill for the single-heuristic setting. The frontmatter declares the design setting, applicable conditions, and activation triggers; the body fixes the unit interface, the design knowledge, the failure modes, and the acceptance criteria.}
\label{lst:skill-paradigm}
\end{figure}

\paragraph{Method skill.}
A method skill carries the methodological knowledge of an established algorithm and is supplied to the agent without importing its native search loop. Listing~\ref{lst:skill-method} shows the EoH method skill, whose body states the operator taxonomy, the recommended parameters, and the execution discipline that the agent follows.

\begin{figure}[!ht]
\centering
\begin{codebox}
\begin{lstlisting}[style=boxed]
---
name: eoh
version: 1.2.0
(*@\textbf{paradigm: evolutionary}@*)
triggers: [eoh, evolution of heuristics]
features:
  n_objectives: 1
  n_units: 1
  unit_kinds: [function]
  domain: [tsp]
---

(*@\textbf{\# Evolution of Heuristics (EoH) Skill}@*)

(*@\textbf{\#\# Operator Taxonomy}@*)
- (*@\textbf{E1 (explore)}@*): mutate an operator into a new form
- (*@\textbf{E2 (exploit)}@*): recombine the strongest operators
- (*@\textbf{M1 (merge)}@*): combine two operators into one
- (*@\textbf{M2 (modify)}@*): tune the parameters of an operator

(*@\textbf{\#\# Parameters}@*)
pop_size = 4, selection_num = 2, max_generations = 10

(*@\textbf{\#\# Execution Guide}@*)
- Emit one operator per turn and record which one was used
- Maintain an implicit population of the best combinations
- Rotate E1 -> E2 -> M1 -> M2 to balance exploration
- On stagnation, prefer E1 over M2

(*@\textbf{\#\# Experience Upgraded}@*)
- Validated on cvrp_dr (swing ~0.145)
- Validated on fjsp_multi_operator (swing ~18.0)
\end{lstlisting}
\end{codebox}
\caption{A method skill. The frontmatter declares the methodology, applicable conditions, and activation triggers; the body states the operator taxonomy, parameters, execution guide, and the accumulated cross-task evidence from skill evolution.}
\label{lst:skill-method}
\end{figure}

\paragraph{Experience card.}
Listing~\ref{lst:card-example} shows a validated card created when an evaluation improved the incumbent. Each line corresponds to a field of Table~\ref{tab:experience-card}: \texttt{problem} and \texttt{situation} fix the context; \texttt{unit} and \texttt{skill} identify what was modified and under which skill; \texttt{content} records the design pattern; \texttt{evidence} and \texttt{metrics} summarize its empirical outcome; and \texttt{code} stores a reference implementation that can be injected into the agent's context.

\begin{figure}[!ht]
\centering
\begin{codebox}
\begin{lstlisting}[style=boxed]
(*@\textbf{problem:}@*)   TSP
(*@\textbf{situation:}@*) stagnation
(*@\textbf{mode:}@*)      validated
(*@\textbf{unit:}@*)      select_next_node
(*@\textbf{skill:}@*)     eoh
(*@\textbf{content:}@*)   Replace the distance-only score with a regret-based
           score that also rewards isolated nodes (large gap to the
           nearest unvisited node), escaping the greedy optimum.
(*@\textbf{evidence:}@*)  9 submissions improved, swing ~0.711
(*@\textbf{code:}@*)      def select_next_node(current_node, destination_node,
                                  unvisited_nodes, distance_matrix):
               regret = [min(distance_matrix[j][k]
                             for k in unvisited_nodes if k != j)
                         for j in unvisited_nodes]
               score = [distance_matrix[current_node][j] - 0.3 * regret[i]
                        for i, j in enumerate(unvisited_nodes)]
               return unvisited_nodes[argmin(score)]
(*@\textbf{metrics:}@*)   {best_score: 6.362, eu: 0.29, situations: [stagnation]}
\end{lstlisting}
\end{codebox}
\caption{A validated experience card from a TSP run. The card records the situation, the modified unit, the design pattern, the empirical evidence, a reference implementation, and self-measured metrics.}
\label{lst:card-example}
\end{figure}

\FloatBarrier
\paragraph{Prompt design.}
\label{app:method-prompt}

The full discovery pipeline is injected into the agent through a single system prompt that states the method and the current task, while each subtask the agent performs is additionally guided by a dedicated reference instruction. We present both below.

\paragraph{System prompt.}
The system prompt establishes the agent's role, describes the closed-loop method, and provides the concrete task; it is followed by the active skill, the retrieved experience cards, and the knowledge state. Listing~\ref{lst:system-prompt} shows a complete example.

\begin{prompttemplate}
\begin{lstlisting}[style=boxed]
You are AlgoEvo, an autonomous agent that designs
high-quality algorithms for optimization problems.

(*@\textbf{\#\# METHOD}@*)
You operate in a closed-loop, experience-driven process:
(*@\textbf{1. Activate.}@*) A design skill is activated for this task;
   it fixes the paradigm, the unit roles, and the guidance.
(*@\textbf{2. Match.}@*) An experience tree indexed by (skill, problem,
   evaluator) is matched; if none exists, create one.
(*@\textbf{3. Search.}@*) You iteratively act on the algorithm: read
   code, edit a unit, diagnose, or evaluate a candidate,
   choosing each action yourself from the current state.
(*@\textbf{4. Summarize.}@*) Every evaluation produces an experience
   card recording the situation, the modification, the
   reward, and the rationale.
(*@\textbf{5. Evolve.}@*) Across tasks, the accumulated experience trees
   are summarized and the skill is refined.

(*@\textbf{\#\# TASK}@*)
(*@\textbf{PROBLEM:}@*) TSP
(*@\textbf{OBJECTIVES:}@*) tour length (minimize)
(*@\textbf{EVOLVABLE UNITS:}@*)
- select_next_node (heuristic) file=construct.py
  signature=def select_next_node(...) -> int
(*@\textbf{BUDGET:}@*) max_samples=100, max_evals=500

(*@\textbf{\#\# DESIGN PRINCIPLES}@*)
1. The code you write MUST be callable by the eval entry.
2. Only modify the internal implementation of the units.
3. Keep the public signatures unchanged.
4. Every submitted candidate must be a runnable file.

(*@\textbf{\#\# ACTIVE SKILL}@*)
===== SKILL: eoh (paradigm: population-based) =====
<design principles, operator taxonomy, failure modes>

(*@\textbf{\#\# EXPERIENCE}@*)
PREVIOUS EXPERIENCE ON THIS PROBLEM (TSP):
STRATEGIES THAT WORKED:
  [EXP-0] select_next_node: weighted distance scoring
    REFERENCE CODE:
    def select_next_node(current_node, destination_node,
                         unvisited_nodes, distance_matrix):
        scores = [distance_matrix[current_node][j] * (1 + angle(j))
                  for j in unvisited_nodes]
        return unvisited_nodes[argmin(scores)]
STRATEGIES TO AVOID:
  [EXP-1] select_next_node: random selection (failed x5)

(*@\textbf{\#\# KNOWLEDGE STATE}@*)
- Step 1: tried greedy heuristic, got 7.231
- Step 2: added 2-opt, improved to 6.892
\end{lstlisting}
\end{prompttemplate}
\vspace{-4pt}
\begin{center}
\small\textit{System prompt example: the method and the task are stated explicitly, followed by the active skill, the retrieved experience cards, and the knowledge state.}
\end{center}
\label{lst:system-prompt}

\paragraph{Subtask instructions.}
For each subtask the agent performs autonomously, a reference instruction specifies its objective, its inputs, and the expected output. Listing~\ref{lst:subtask-prompts} summarizes the four instructions: skill activation, experience tree matching, experience card summarization, and skill refinement.

\begin{instructiontemplate}
\begin{lstlisting}[style=boxed]
(*@\textbf{[SKILL ACTIVATION]}@*)
Given the problem description and objectives, select the
single most appropriate design skill from the registry
and justify the choice.

(*@\textbf{[EXPERIENCE TREE MATCHING]}@*)
Search the experience bank for a tree indexed by
(skill, problem, evaluator) that matches the current
task. If one exists, load it; otherwise create a new
empty tree. Report the matched or created tree.

(*@\textbf{[EXPERIENCE CARD SUMMARIZATION]}@*)
Summarize the latest evaluation into an experience card
with the fields: problem, situation, mode (validated |
avoid | reflection), unit, skill, content (the design
pattern), evidence (the empirical outcome), and
optional reference code and metrics.

(*@\textbf{[SKILL REFINEMENT]}@*)
Review the experience trees owned by the skill. Summarize
the most salient recurring patterns and refine the skill
document with them, without overwriting its
paradigm-level guidance.
\end{lstlisting}
\end{instructiontemplate}
\label{lst:subtask-prompts}

\section{Benchmark and Evaluation Details}
\label{app:setup}

\subsection{Benchmark Tasks and Instance Sets}
\label{app:tasks}
To comprehensively evaluate \systemname{} across distinct algorithmic complexities, we utilize six benchmark tasks spanning single-heuristic construction, multi-objective optimization, and multi-component co-design. Each task incorporates a training set utilized during the discovery process and a disjoint test set to evaluate generalization. All comparative methods operate under identical instance splits. Table~\ref{tab:tasks} summarizes the instance distributions and objectives.

\begin{table}[!ht]
\centering
\footnotesize
\setlength{\tabcolsep}{2pt}
\caption{Summary of benchmark tasks, objectives, and instance distributions.}
\label{tab:tasks}
\begin{tabular}{@{}l>{\raggedright\arraybackslash}p{0.22\linewidth}>{\raggedright\arraybackslash}p{0.32\linewidth}>{\raggedright\arraybackslash}p{0.32\linewidth}@{}}
\toprule
\textbf{Task} & \textbf{Objective} & \textbf{Training set} & \textbf{Test set} \\
\midrule
\cellcolor{grpA}TSP
& tour length $\downarrow$
& 16 instances (50 cities, uniform)
& 24 instances (sizes 50/100/200, uniform/clustered), plus 6 TSPLib \\
\cellcolor{grpA}CVRP
& route cost $\downarrow$
& 16 instances (50 customers, cap 40, demands $U[1,10]$)
& 48 instances (16 each of sizes 50, 100, and 200) \\
\midrule
\cellcolor{grpB}Bi-TSP
& two tour lengths $\downarrow$
& 1 instance (50 cities, two coordinate matrices)
& 2 instances (sizes 20 and 100) \\
\cellcolor{grpB}Bi-FJSP
& makespan $\downarrow$, utilization $\uparrow$
& 10 Brandimarte instances (mk01--mk10)
& 5 Brandimarte instances (mk11--mk15) \\
\midrule
\cellcolor{grpC}CVRP-DR
& route cost $\downarrow$
& 10 instances (50 customers, cap 50, demands $[0,9]$)
& 64 instances (size 50), plus 5 reserve sets up to size 500 \\
\cellcolor{grpC}FJSP 4-Ops
& makespan $\downarrow$
& 10 Brandimarte instances (mk01--mk10)
& 5 Brandimarte instances (mk11--mk15) \\
\bottomrule
\end{tabular}
\end{table}

\subsection{Metrics}
\label{app:metrics}
For single-objective tasks, performance is the mean objective over training instances, $\bar{F}_d(\mathbf{\Pi})=\frac{1}{|\mathcal{D}_{\mathrm{train}}|}\sum_{\mathbf{x}\in\mathcal{D}_{\mathrm{train}}}F_d(\mathbf{x}\mid\mathbf{\Pi})$, where lower is better. For multi-objective tasks, non-dominated objective vectors form an archive, and objectives are min-max normalized per problem, $\tilde{z}_i=(z_i-z_i^{\min})/(z_i^{\max}-z_i^{\min})$, over all solutions evaluated across methods and seeds. We report HV~\citep{zitzler1998multiobjective} and IGD~\citep{coello2005solving}, with reference point $\mathbf{r}=(1.1,\ldots,1.1)$; HV is the Lebesgue measure $\Lambda$ of the region dominated by the archive $\mathcal{P}$,

\begin{equation}
\mathrm{HV}(\mathcal{P})
=
\frac{1}{\prod_{i=1}^{M}r_i}
\Lambda
\left(
\bigcup_{\mathbf{z}\in\mathcal{P}}
[\mathbf{z},\mathbf{r}]
\right),
\label{eq:app-hv}
\end{equation}

and IGD is the mean distance from the reference front $\mathcal{F}$ of the aggregated solutions to $\mathcal{P}$,

\begin{equation}
\mathrm{IGD}(\mathcal{P})
=
\frac{1}{|\mathcal{F}|}
\sum_{\mathbf{u}\in\mathcal{F}}
\min_{\mathbf{z}\in\mathcal{P}}
\lVert\mathbf{u}-\mathbf{z}\rVert_2,
\label{eq:app-igd}
\end{equation}

with higher HV and lower IGD indicating better convergence and diversity. Both the search reward and the reported metrics use a hypervolume reference point; the reported values are additionally normalized to $[0,1]$ over the union of evaluated solutions so that they are comparable across methods and problems. As normalization bases are built per problem, these metrics are compared within a problem only.

\subsection{Baseline Details}
\label{app:baselines}

All baselines are re-implemented from their original publications and evaluated on the LLM4AD platform~\citep{liu2024llm4ad} under the shared protocol: identical instance pools, the same evaluation entry points, a maximum of $500$ evaluations per seed, and three random seeds.

\paragraph{Rule-based heuristics.}
We include simple constructive rules that involve no search, as classical reference points. For TSP, the nearest-neighbour (NN) heuristic builds a tour by repeatedly moving from the current city $i$ to the closest unvisited city $\arg\min_j d(i,j)$. For CVRP, NN is made capacity-aware: from the current customer it visits the nearest unvisited customer whose demand fits the remaining vehicle capacity, and opens a new route only when no such customer exists. For the multi-component CVRP-DR task, the \emph{Standard} baseline runs the shipped destroy-and-repair pipeline, whose three units are a savings-inspired edge score (inverse distance weighted by capacity efficiency and a depot bonus), a worst-customer removal criterion (demand-to-capacity ratio plus mean distance to other customers), and a nearest-insertion repair rule. For FJSP 4-Ops, the \emph{Expert} baseline uses the hand-designed operator set shipped with the task. These rules are the starting points that every search-based method must improve upon.

\paragraph{Search-based baselines.}
For single-heuristic construction we compare against EoH~\citep{eoh2024}, ReEvo~\citep{reevo2024}, MCTS-AHD~\citep{mctsahd2025}, and FunSearch~\citep{funsearch2024}, each following its original search loop. For multi-objective design we use MEoH~\citep{meoh2025}, NSGA-II~\citep{deb2002nsga2}, and MOEA/D~\citep{zhang2007moead}, whose archives pass through the shared normalization and metric harness. For multi-component co-design we include MOTIF~\citep{motif2025} and E2OC~\citep{e2oc2026}, together with the two EoH variants described next.

\paragraph{Component-selection variants.}
Synergy and Rotating-EoH adapt EoH to multi-component tasks by deciding which component to edit at each step, and they differ precisely in that decision. \emph{Synergy} keeps an EoH population over complete component combinations and, at each iteration, draws the component to edit at random with a probability weighted by stagnation, so that a component whose recent evaluations have failed to improve the baseline receives a higher weight and effort is redirected toward the weakest component. \emph{Rotating-EoH} instead rotates through the components in a fixed order, spending a small candidate budget on each component in turn and keeping a change only if it improves the incumbent, which amounts to greedy coordinate descent over the components. Both use the same EoH operators and the shared evaluation harness, and neither measures cross-component coupling, which is the signal \systemname{} exploits through its diagnostic probes.

\paragraph{Parameter setting.}
Across all methods the controlled quantity is the sampling budget, i.e.\ the maximum number of candidate evaluations; each search-based baseline otherwise keeps the hyperparameters reported in its original publication, such as population size, archive capacity, and its selection or variation operators, so that every method runs in the configuration its authors found effective. The surrounding environment is fixed and follows the LLM4AD benchmark~\citep{liu2024llm4ad}: identical instance pools and splits, the same evaluation entry points, the same initial code templates for each evolvable unit, and the same feasibility and objective routines. The comparison therefore isolates the search procedure rather than environment-specific tuning.

\subsection{Anti-cheating settings}
\label{app:anti-cheat}

Because the agent writes and runs its own code, each run is guarded so that a reported result reflects genuine algorithm design rather than a shortcut around the intended search.

\paragraph{Workflow control.}
Each run is driven by the system prompt and subtask reference instructions of Appendix~\ref{app:method-prompt}, which specify the task, the evolvable units, and how the design skill hub (Section~\ref{sec:hub}) and the experience bank (Section~\ref{sec:expbank}) are used at each stage of Algorithm~\ref{alg:algoevo}. The agent records every action and its outcome in the working context, so the trajectory is auditable and the task is carried to completion.

\paragraph{Exception handling.}
Every step is checked against its returned result and the log of tool invocations. A step that raises an execution error, returns a malformed result, or yields an invalid candidate is discarded, recorded in the working context, and re-executed, so that a transient failure is neither counted as progress nor allowed to erase the accumulated context.

\paragraph{Budget integrity.}
A run could otherwise score candidates through a private script and evade the budget $B$ of Appendix~\ref{app:method-task}, which counts only calls to the shared evaluation entry point. Counters are therefore installed at both the instance layer and the evaluation entry point: a candidate is counted only when it is evaluated through the evaluation action of Table~\ref{tab:actions-app}, and the counter is cross-checked against the number of individuals the algorithm generated. A mismatch, or any direct call to the objective that bypasses the evaluator, invalidates the run, which is discarded rather than reported.

\paragraph{Isolation.}
Runs execute in isolated sandboxes with separate working directories and process spaces, so that files and state cannot leak across runs or between methods.

% ============================================================
% E. Design Space and Task Instantiation
% ============================================================

\section{Design Space and Task Instantiation}
\label{app:units}

To isolate the contribution of agentic algorithm discovery, \systemname{} keeps the surrounding solver, instance handling, feasibility checks, and objective evaluation fixed, restricting optimization exclusively to pre-defined evolvable algorithmic units. Table~\ref{tab:units-summary} summarizes the design space configuration across the six benchmark tasks.

\begin{table}[!ht]
\centering
\footnotesize
\setlength{\tabcolsep}{5pt}
\caption{Evolvable algorithmic units across the six benchmark tasks.}
\label{tab:units-summary}
\begin{tabular}{@{}lll@{}}
\toprule
\textbf{Task} & \textbf{Number of units} & \textbf{Designed components} \\
\midrule
\cellcolor{grpA}TSP & 1 & Next-city selection rule \\
\cellcolor{grpA}CVRP & 1 & Capacity-aware next-customer rule \\
\cellcolor{grpB}Bi-TSP & 1 & Bi-objective next-city selection rule \\
\cellcolor{grpB}Bi-FJSP & 1 & Dispatching rule \\
\cellcolor{grpC}CVRP-DR & \hpC{3} & Construction, destruction, and repair rules \\
\cellcolor{grpC}FJSP 4-Ops & \hpC{4} & Two crossover and two mutation operators \\
\bottomrule
\end{tabular}
\end{table}

\paragraph{Single-Unit Design Tasks.}
\label{app:single-unit}

For TSP, the designed component is a scoring rule that selects the next unvisited city. The surrounding constructive execution and termination logic remain fixed.

\begin{unitbox}
\noindent\textbf{Evolvable Unit: Next-Node Scoring Function}\par\medskip
\footnotesize
\begin{verbatim}
def select_next_node(current_node, destination_node,
                     unvisited_nodes, distance_matrix) -> int
\end{verbatim}
\end{unitbox}

For CVRP, the designed component governs capacity-aware customer selection and depot-return timing. The fixed evaluator tracks remaining vehicle load constraints and filters out infeasible choices.

\begin{unitbox}
\noindent\textbf{Evolvable Unit: Capacity-Aware Next-Node Selection}\par\medskip
\footnotesize
\begin{verbatim}
def select_next_node(current_node, depot, unvisited_nodes,
                     rest_capacity, demands, distance_matrix) -> int
\end{verbatim}
\end{unitbox}

For Bi-TSP, the same constructive rule is evaluated across two distinct distance objectives simultaneously, enabling the generation of Pareto archives through multi-objective feedback.

\begin{unitbox}
\noindent\textbf{Evolvable Unit: Bi-Objective Next-City Selection}\par\medskip
\footnotesize
\begin{verbatim}
def select_next_city(current_node, unvisited_nodes,
                     dist_1, dist_2) -> int
\end{verbatim}
\end{unitbox}

For Bi-FJSP, the designed component acts as a priority dispatching rule determining task execution order at each scheduling step. The fixed evaluator enforces precedence constraints and machine eligibility.

\begin{unitbox}
\noindent\textbf{Evolvable Unit: Operation Dispatching Rule}\par\medskip
\footnotesize
\begin{verbatim}
def select_operation(job_status, machine_status, feasible_ops) -> tuple
\end{verbatim}
\end{unitbox}

\paragraph{CVRP-DR: Sequential Component Coupling.}
\label{app:cvrpdr}

CVRP-DR instantiates a multi-component design space comprising three sequential stages: construction, destruction, and repair. Algorithm~\ref{alg:dr} outlines the evaluation procedure.

\begin{unitbox}
\noindent\textbf{Evolvable Units: Destroy-and-Repair Components (F1/F2/F3)}\par\medskip
\footnotesize
\begin{verbatim}
def edge_score(i, j, distances, demands, capacity) -> float
def customer_badness(customer_idx, permutation, distances,
                     demands, capacity) -> float
def insert_position(customer, permutation, distances,
                    demands, capacity) -> int
\end{verbatim}
\end{unitbox}

\begin{algorithm}[h]
\caption{CVRP-DR evaluation procedure}
\label{alg:dr}
\begin{algorithmic}[1]
\Require Instance, construction rule $\Pi_c$, destruction rule $\Pi_d$, repair rule $\Pi_r$, removal fraction $\rho$
\State \hlrow{Construct an initial solution using $\Pi_c$}
\State \hlrow{Evaluate customer badness using $\Pi_d$}
\State Remove the $\lceil \rho n\rceil$ customers with the highest badness scores
\For{each removed customer $c$}
\State \hlrow{Select an insertion position using $\Pi_r$}
\State Insert $c$ into the selected position
\EndFor
\State Apply the fixed local-improvement procedure
\State \Return Solution cost
\end{algorithmic}
\end{algorithm}

These three components exhibit sequential coupling: the output of construction bounds the scope of destruction, and the partial state left by destruction directly dictates the restoration target for repair. Consequently, optimizing these strategies requires coordinated reasoning rather than isolated component tuning.

\paragraph{FJSP 4-Ops: Representation-Based Coupling.}
\label{app:fjsp4ops}

FJSP 4-Ops implements a multi-component genetic programming structure that jointly optimizes four distinct operators: priority-segment crossover, priority-segment mutation, machine-assignment crossover, and machine-assignment mutation. Chromosomes are structured as concatenated segments:

\begin{equation}
\mathbf{c}
=
[\mathbf{c}^{\mathrm{priority}}
\mid
\mathbf{c}^{\mathrm{machine}}].
\label{eq:chromosome}
\end{equation}

The priority segment determines operation sequences within jobs, while the machine segment dictates resource assignments. Algorithm~\ref{alg:ga} details the simplified evaluation loop.

\begin{unitbox}
\noindent\textbf{Evolvable Units: Genetic Operators (op/ma $\times$ crossover/mutation)}\par\medskip
\footnotesize
\begin{verbatim}
def op_crossover(parent1, parent2, n_vars) -> (child1, child2)
def op_mutation(solution, n_vars) -> solution
def ma_crossover(parent1, parent2, n_vars) -> (child1, child2)
def ma_mutation(solution, n_vars) -> solution
\end{verbatim}
\end{unitbox}

\begin{algorithm}[h]
\caption{FJSP four-operator genetic algorithm evaluation}
\label{alg:ga}
\begin{algorithmic}[1]
\Require Population and four evolvable operators
\State Initialize chromosomes
\For{each generation}
\State Evaluate population fitness and retain elite solutions
\While{population is not refilled}
\State Select parent chromosomes
\State \hlrow{Apply priority crossover with probability $0.5$}
\State \hlrow{Apply priority mutation with probability $0.3$}
\State \hlrow{Apply machine crossover with probability $0.5$}
\State \hlrow{Apply machine mutation with probability $0.2$}
\State Legalize and evaluate offspring
\EndWhile
\EndFor
\State \Return Best makespan
\end{algorithmic}
\end{algorithm}

Unlike the pipeline coupling in CVRP-DR, FJSP 4-Ops couples components through a shared chromosomal representation and a joint decoding mechanism. Together, these multi-component tasks rigorously evaluate the framework's ability to coordinate interdependent strategies across diverse structural paradigms.

% ============================================================
% F. Additional Experimental Results
% ============================================================

\section{Additional Experimental Results}
\label{app:additional}

\subsection{Convergence Under the Evaluation Budget}
\label{app:convergence}

\begin{figure}[t]
\centering

\begin{subfigure}[t]{0.49\linewidth}
\centering
\includegraphics[width=\linewidth]{figures/fig_qb_tsp50.pdf}
\caption{TSP.}
\label{fig:appbudget-a}
\end{subfigure}
\hfill
\begin{subfigure}[t]{0.49\linewidth}
\centering
\includegraphics[width=\linewidth]{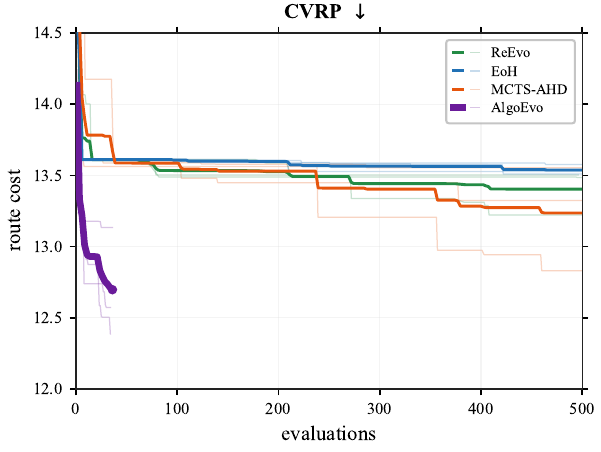}
\caption{CVRP.}
\label{fig:appbudget-b}
\end{subfigure}

\begin{subfigure}[t]{0.49\linewidth}
\centering
\includegraphics[width=\linewidth]{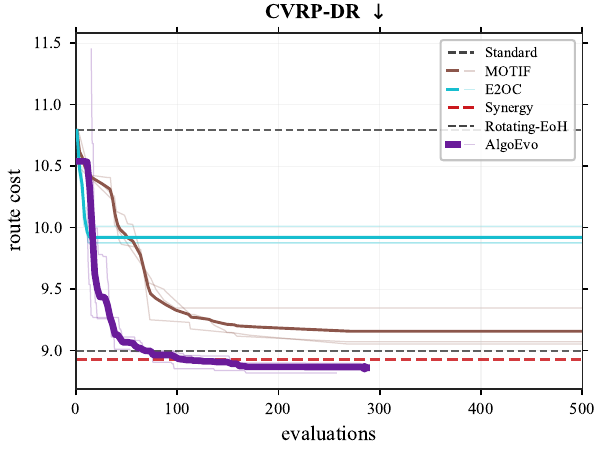}
\caption{CVRP-DR.}
\label{fig:appbudget-c}
\end{subfigure}
\hfill
\begin{subfigure}[t]{0.49\linewidth}
\centering
\includegraphics[width=\linewidth]{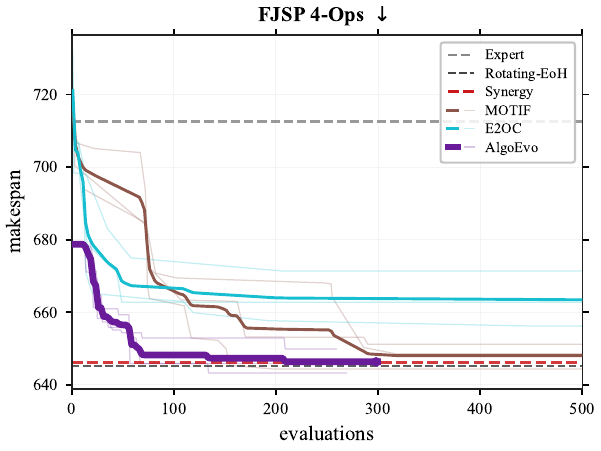}
\caption{FJSP 4-Ops.}
\label{fig:appbudget-d}
\end{subfigure}

\begin{subfigure}[t]{0.49\linewidth}
\centering
\includegraphics[width=\linewidth]{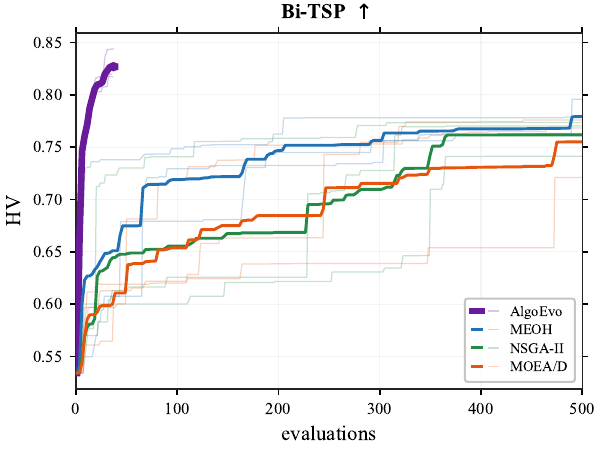}
\caption{Bi-TSP.}
\label{fig:appbudget-e}
\end{subfigure}
\hfill
\begin{subfigure}[t]{0.49\linewidth}
\centering
\includegraphics[width=\linewidth]{figures/fig_qb_fjsp_biobj.pdf}
\caption{Bi-FJSP.}
\label{fig:appbudget-f}
\end{subfigure}

\caption{Best-so-far performance against the evaluation budget across the six design tasks.}
\label{fig:appbudget}
\end{figure}

Figure~\ref{fig:appbudget} illustrates the best-so-far performance as a function of the evaluation budget across all six tasks, complementing the final results in the main text by showing how design quality evolves during the search process.

In single-heuristic and multi-objective tasks, \systemname{} consistently identifies competitive designs within a small fraction of the available budget. Conversely, multi-component tasks demand longer search trajectories, reflecting larger solution spaces and complex interactions among multiple evolvable components. These trajectories demonstrate the framework's capacity for adaptive resource allocation without prioritizing raw evaluation counts.

\subsection{Pareto Archives on Multi-Objective Tasks}
\label{app:pareto}

\begin{figure}[t]
\centering

\begin{subfigure}[t]{0.49\linewidth}
\centering
\includegraphics[width=\linewidth]{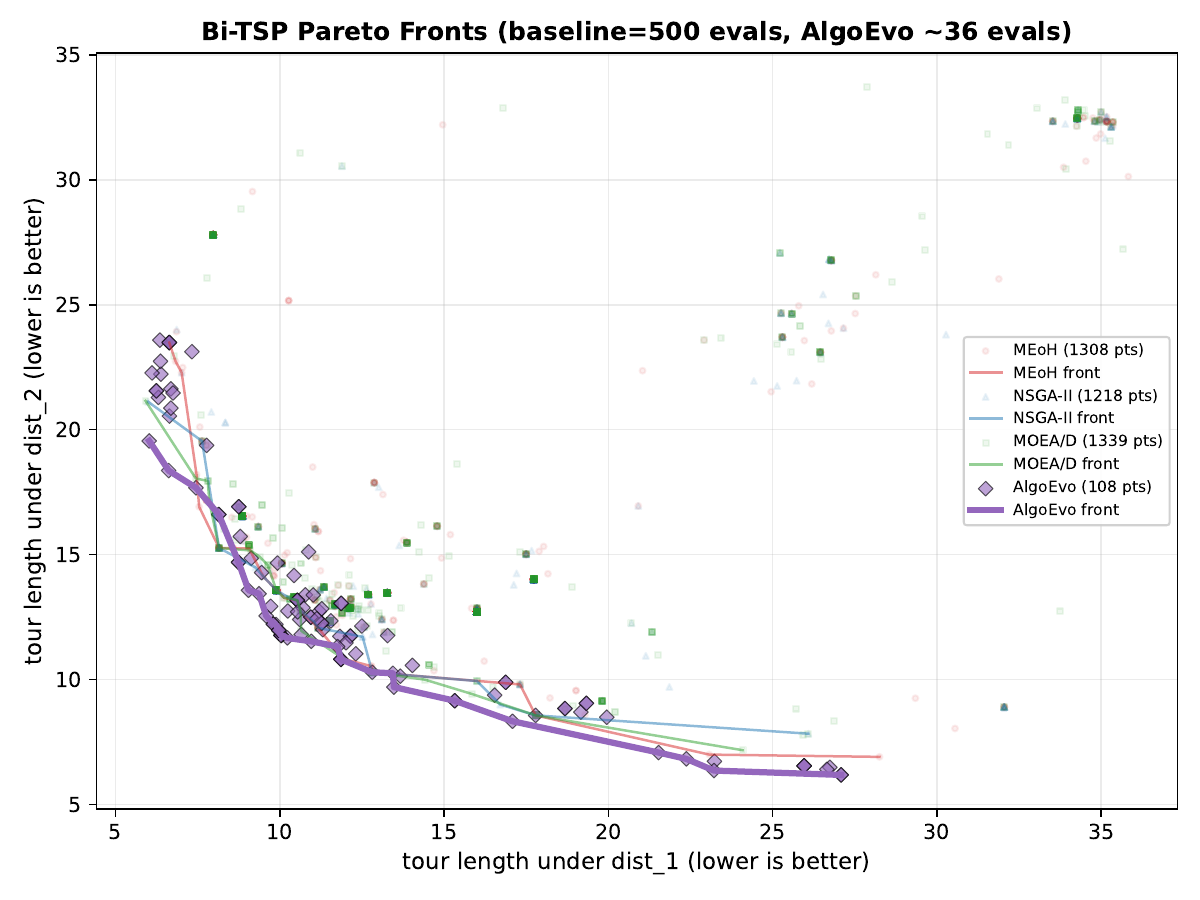}
\caption{Bi-TSP.}
\label{fig:apppareto-tsp}
\end{subfigure}
\hfill
\begin{subfigure}[t]{0.49\linewidth}
\centering
\includegraphics[width=\linewidth]{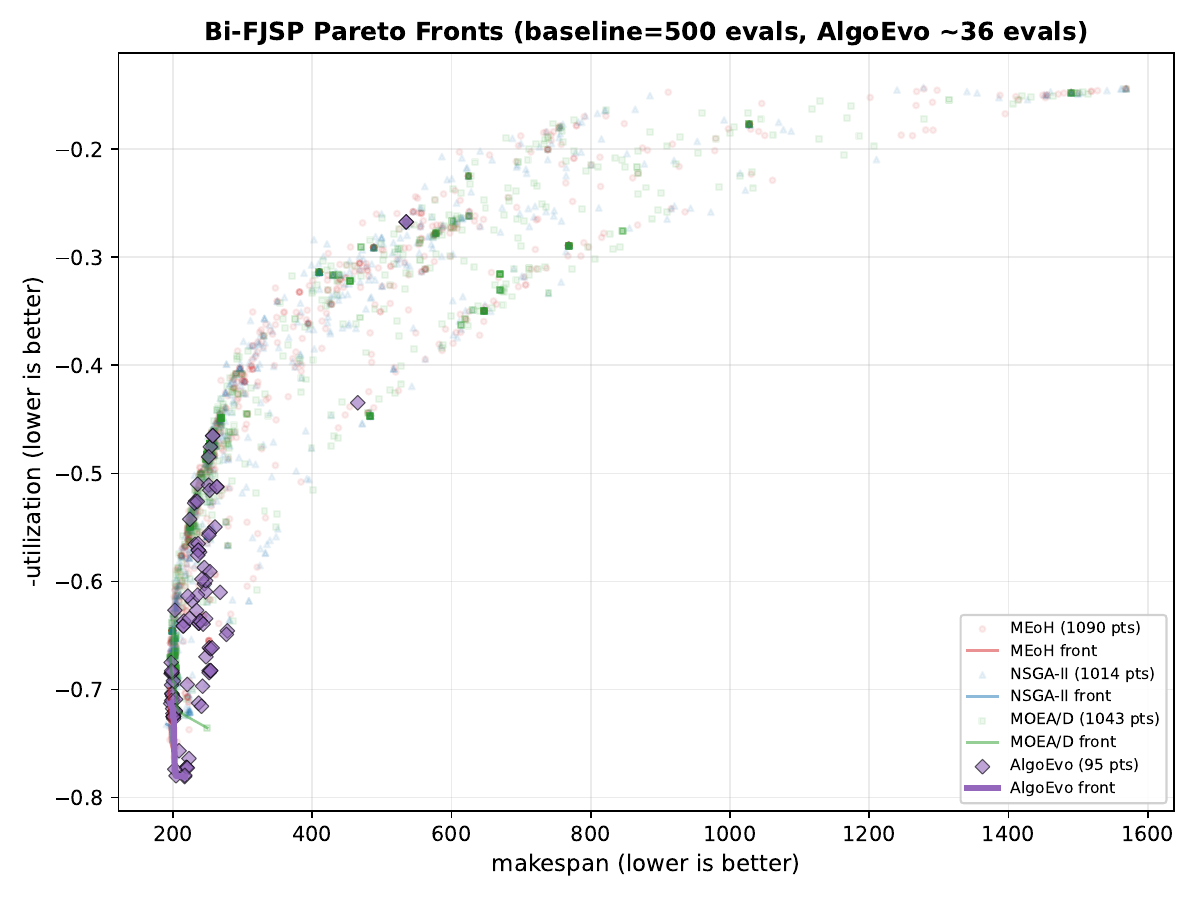}
\caption{Bi-FJSP.}
\label{fig:apppareto-fjsp}
\end{subfigure}

\caption{Pareto archives on multi-objective tasks, displaying all evaluated candidates across three random seeds.}
\label{fig:apppareto}
\end{figure}

Figure~\ref{fig:apppareto} presents the Pareto archives corresponding to the multi-objective experiments. To display the broader distribution of search outcomes alongside final non-dominated sets, evaluated candidates across all three random seeds are included.

The visualized archives align with the aggregate HV and IGD outcomes in the main text. \systemname{} explores a wider region of the objective space and yields non-dominated solutions closer to optimal trade-off frontiers, supporting multi-objective algorithm design where discovering complementary design sets is paramount.

\subsection{Ablation of Knowledge Mechanisms}
\label{app:ablation}

To assess the contributions of core architectural components, we conduct an ablation study on CVRP-DR under standard evaluation protocols, summarized in Table~\ref{tab:ablation-app}. Because this experiment runs as an independent batch, absolute values may vary slightly from the primary benchmark table.

\begin{table}[!ht]
\centering
\footnotesize
\setlength{\tabcolsep}{4pt}
\caption{Ablation results on CVRP-DR over three random seeds.}
\label{tab:ablation-app}
\begin{tabular}{@{}lcc@{}}
\toprule
\textbf{Variant} & \textbf{Cost $\downarrow$} & \textbf{Deviation} \\
\midrule
\systemname{} (complete framework) & $\mathbf{9.040 \pm 0.031}$ & --- \\
without the experience bank & $9.100 \pm 0.132$ & $+0.66\%$ \\
without the design skill hub & $9.092 \pm 0.153$ & $+0.58\%$ \\
without situation-triggered injection & $9.059 \pm 0.008$ & $+0.21\%$ \\
\bottomrule
\end{tabular}
\end{table}

Removing either the experience bank or the design skill hub incurs a larger performance drop than omitting situation-triggered injection. This confirms that persistent memory and reusable design abstractions serve as the primary drivers of discovery performance, whereas situation-triggered injection primarily refines knowledge selection efficiency.
% ============================================================
% G. Further Analysis
% ============================================================

\section{Further Analysis}
\label{app:further-analysis}

\subsection{Method Skills and Methodological Transfer}
\label{app:methodskill}

The design skill hub encodes not only paradigm-level design knowledge but also methodological insights distilled from existing algorithms. To examine whether such knowledge remains effective when decoupled from its original search framework, we construct six method skills from established methods and supply them to \systemname{} without importing their native search procedures.

We evaluate EoH~\citep{eoh2024}, ReEvo~\citep{reevo2024}, and FunSearch~\citep{funsearch2024} for single-heuristic design, and MEoH~\citep{meoh2025}, NSGA-II~\citep{deb2002nsga2}, and MOEA/D~\citep{zhang2007moead} for multi-objective design, ensuring native and skill-based variants share identical seeds and instance sets.

\begin{figure}[h]
\centering
\includegraphics[width=\linewidth]{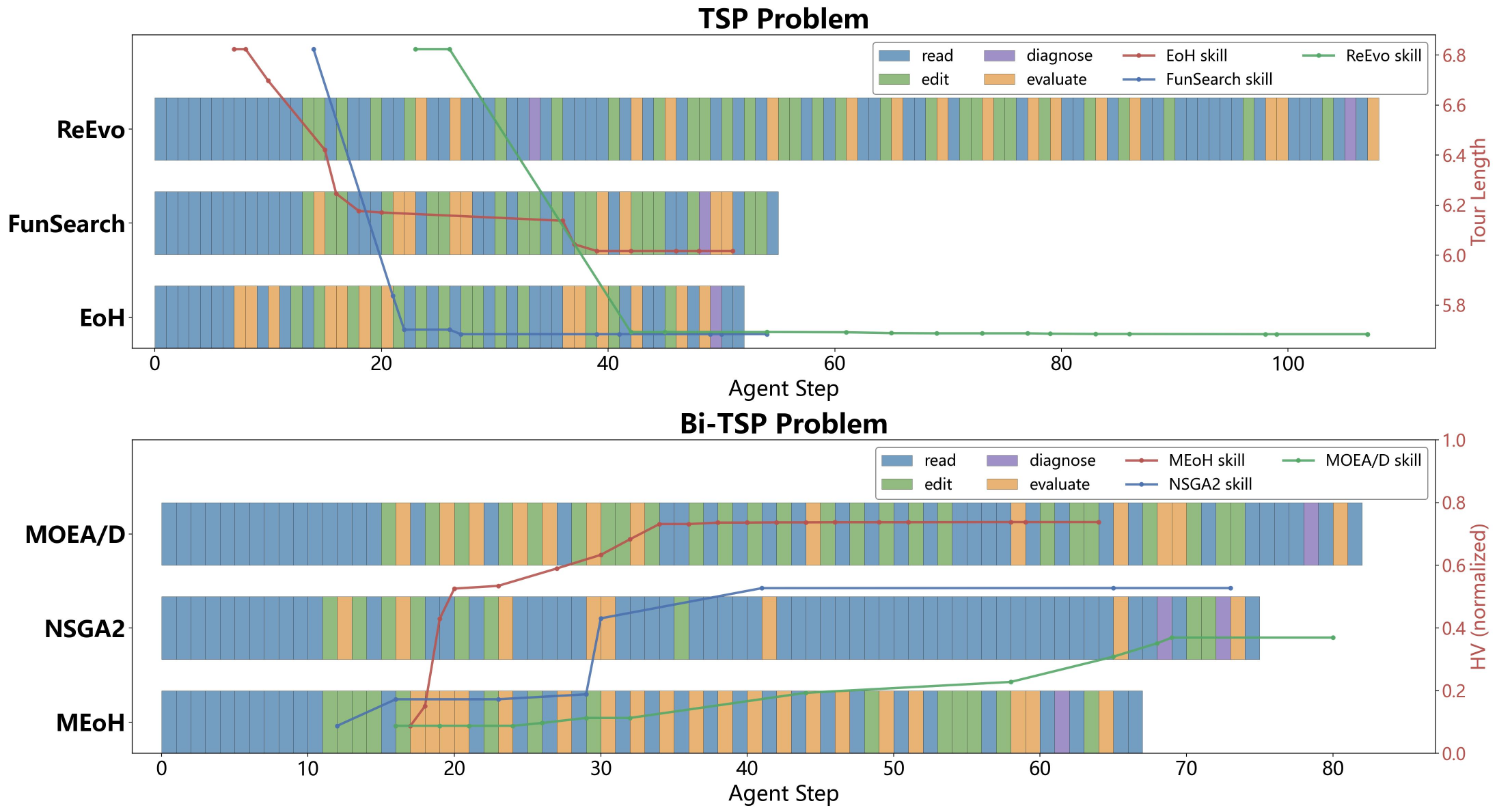}
\caption{Agent behavior and convergence trajectory on TSP (top) and Bi-TSP (bottom). Colored bands indicate the action type of each step, and the curve traces the best-so-far quality against the number of evaluations. The agent follows a structured design process of inspecting, implementing, and revising in response to feedback, rather than random trial-and-error.}
\label{fig:skill_behavior}
\end{figure}

To characterize how the agent behaves when executing a transferred methodology, Figure~\ref{fig:skill_behavior} traces a complete run on TSP (top) and Bi-TSP (bottom). Colored bands mark the action type of each step, and the curve traces the best-so-far quality against the number of evaluations. The agent does not repeatedly rewrite code at random: it first inspects the implementation to understand the problem structure, then produces a candidate, and thereafter conditions each edit on the feedback of the previous evaluation. The trajectory therefore resembles a paced design process of inspection, implementation, evaluation, and revision, rather than trial-and-error, indicating that the agent follows the methodological workflow encoded in the skill instead of merely emitting code.

\begin{table}[!ht]
\centering
\footnotesize
\setlength{\tabcolsep}{4pt}
\caption{Comparison between native methods and \systemname{} equipped with corresponding method skills.}
\label{tab:methodskill}
\begin{tabular}{@{}lcccc@{}}
\toprule
& \multicolumn{2}{c}{\cellcolor{grpA}\textbf{Quality}} & \multicolumn{2}{c}{\cellcolor{grpC}\textbf{Evaluations}} \\
\cmidrule(lr){2-3} \cmidrule(lr){4-5}
\textbf{Method skill} & \textbf{Native} & \textbf{With skill} & \textbf{Native} & \systemname{} \\
\midrule
\multicolumn{5}{l}{\textit{TSP: tour length $\downarrow$}} \\
EoH & $7.1529 \pm 0.0263$ & $\mathbf{7.0414 \pm 0.0923}$ & $500$ & $\mathbf{22}$ \\
ReEvo & $7.3478 \pm 0.3112$ & $\mathbf{7.1889 \pm 0.3678}$ & $500$ & $\mathbf{19}$ \\
FunSearch & $7.6111 \pm 0.1377$ & $\mathbf{7.1815 \pm 0.1057}$ & $500$ & $\mathbf{21}$ \\
\midrule
\multicolumn{5}{l}{\textit{Bi-TSP: HV $\uparrow$}} \\
MEoH & $0.760 \pm 0.025$ & $\mathbf{0.829 \pm 0.004}$ & $500$ & $\mathbf{316}$ \\
NSGA-II & $0.614 \pm 0.082$ & $\mathbf{0.828 \pm 0.006}$ & $500$ & $\mathbf{391}$ \\
MOEA/D & $0.314 \pm 0.221$ & $\mathbf{0.829 \pm 0.006}$ & $500$ & $\mathbf{284}$ \\
\bottomrule
\end{tabular}
\end{table}

\begin{figure}[h]
\centering
\includegraphics[width=0.9\linewidth]{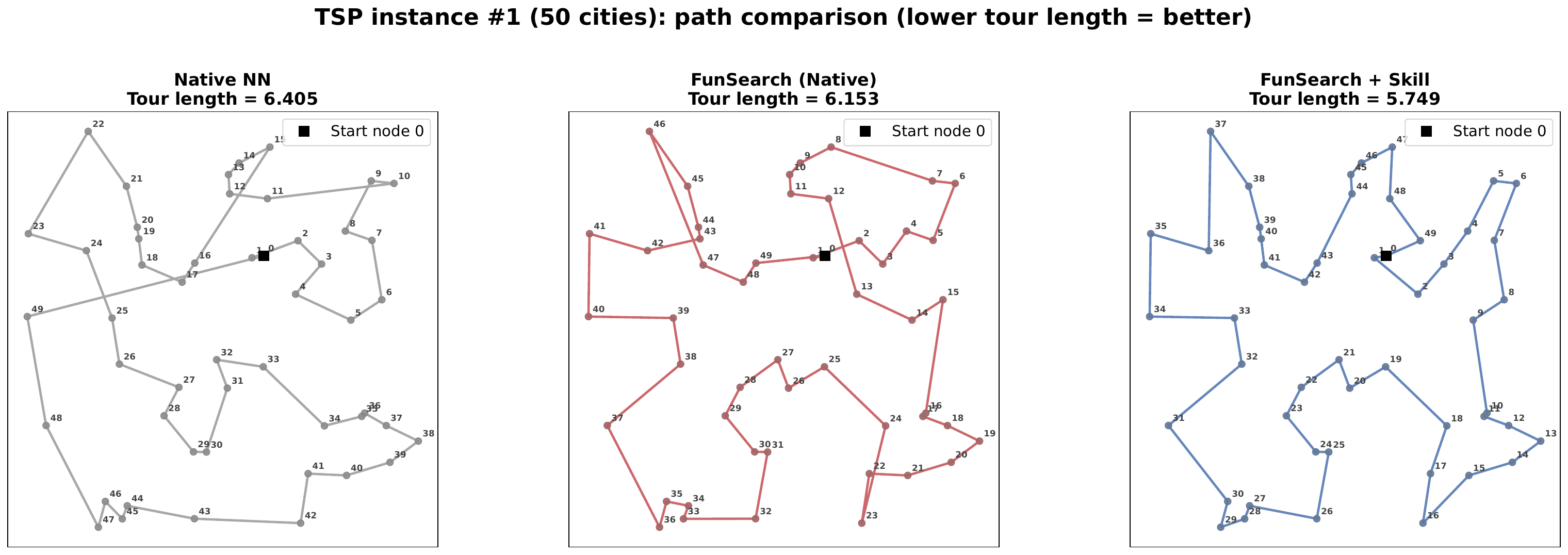}
\caption{TSP path visualization on a 50-city instance. Left: native nearest-neighbor heuristic (tour length 6.405); center: FunSearch native method (6.153); right: \systemname{} with FunSearch skill (5.749). The skill-based variant discovers a more efficient tour structure while using fewer evaluations.}
\label{fig:tsp_path_comparison}
\end{figure}

\begin{figure}[h]
\centering
\includegraphics[width=\linewidth]{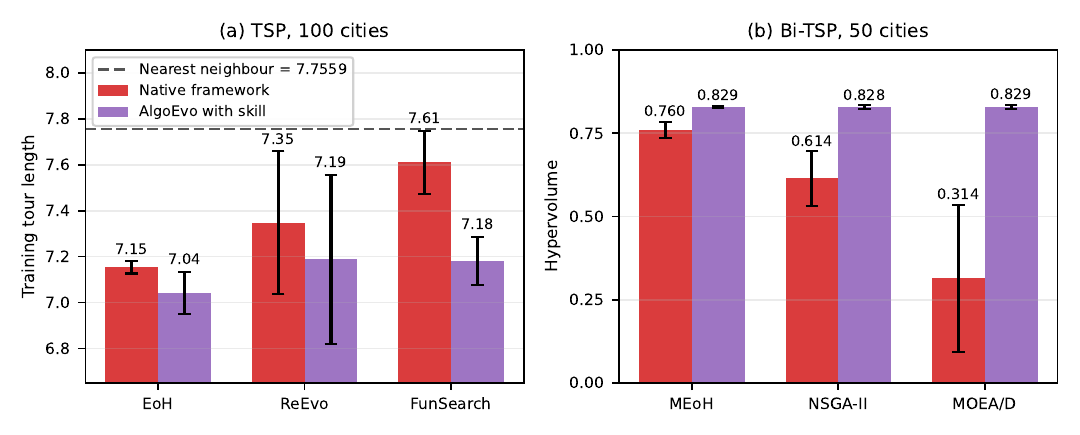}
\caption{Per-seed quality comparison between native methods and \systemname{} with corresponding method skills.}
\label{fig:methodskill}
\end{figure}

Table~\ref{tab:methodskill} reports that the skill-based versions achieve solution quality comparable to or exceeding their native counterparts. On TSP, transferred methodologies reach superior average performance with substantially fewer evaluations, illustrated in Figure~\ref{fig:tsp_path_comparison}. On Bi-TSP, the primary benefit is enhanced archive quality and stability, while evaluation reductions are less pronounced. These findings indicate that methodological knowledge can be effectively represented and utilized independently of its original search framework. Figure~\ref{fig:methodskill} presents the per-seed comparison.

Note that the two experimental blocks use different problem scales (100-city instances for TSP versus 50-city instances for Bi-TSP), and absolute values should not be compared across tasks.

\begin{figure}[h]
\centering
\includegraphics[width=\linewidth]{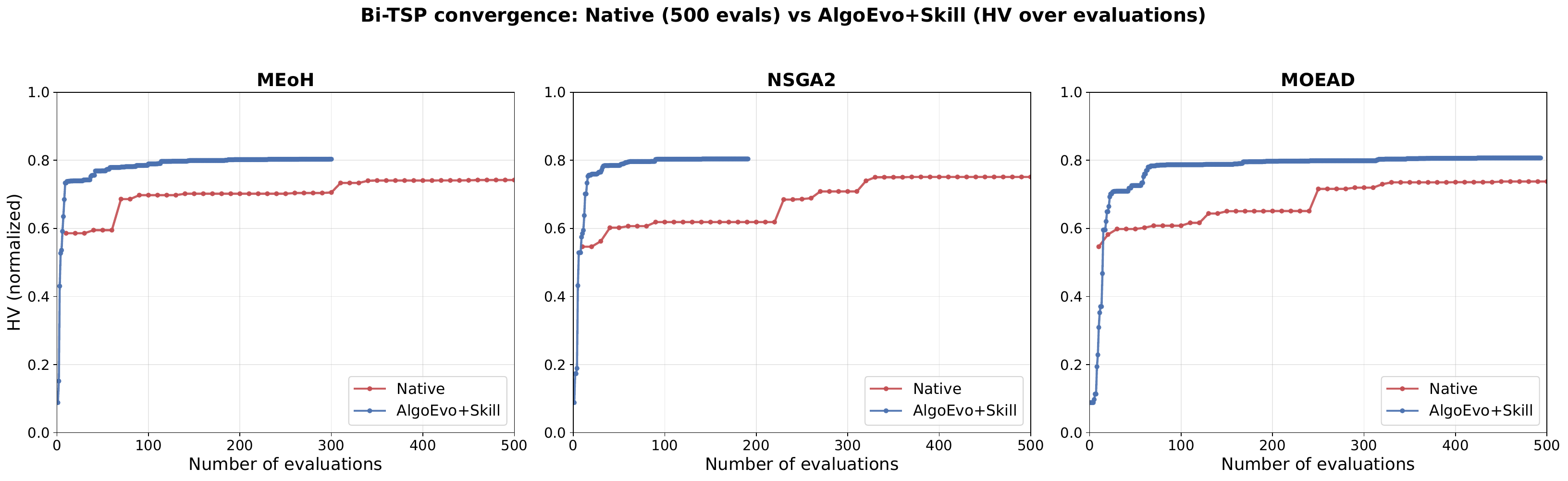}
\caption{Bi-TSP convergence curves: native methods (red, 500 evaluations) versus \systemname{} with method skills (blue). AlgoEvo+Skill achieves higher HV within significantly fewer evaluations across all three methods.}
\label{fig:bitsp_convergence}
\end{figure}

\begin{figure}[h]
\centering
\includegraphics[width=\linewidth]{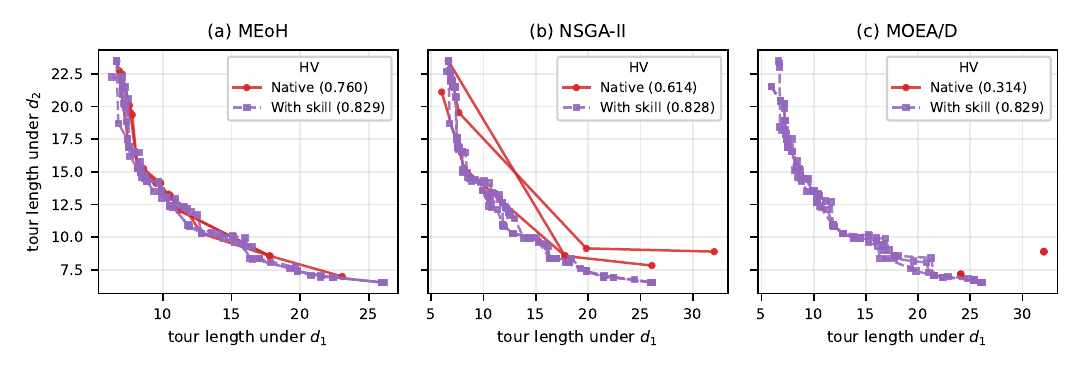}
\caption{Bi-TSP Pareto fronts for native methods and \systemname{} equipped with corresponding method skills.}
\label{fig:methodskillpareto}
\end{figure}

For Bi-TSP, the corresponding Pareto fronts in Figure~\ref{fig:methodskillpareto} show that skill-based runs cover a broader favorable region of the objective space, consistent with their higher HV values and with the convergence curves in Figure~\ref{fig:bitsp_convergence}.

\subsection{Representative Diagnosis-Guided Design Trajectory}
\label{app:trajectory}

To illustrate how diagnostic feedback adjusts the granularity of multi-component search, Table~\ref{tab:trajcase} details a representative CVRP-DR trajectory.

\begin{table}[!ht]
\centering
\footnotesize
\setlength{\tabcolsep}{4pt}
\caption{Representative diagnosis-guided design trajectory on CVRP-DR.}
\label{tab:trajcase}
\begin{tabular}{@{}cllc@{}}
\toprule
\textbf{Step} & \textbf{Operation} & \textbf{Observation} & \textbf{Cost} \\
\midrule
1 & Diagnose & The destroyer exhibits the largest marginal effect & --- \\
2 & Analyse interaction & Destroyer and repairer display strong interaction & --- \\
3 & Edit & Replace the destroyer's distance-based criterion with absorption cost & $9.310$ \\
4 & Diagnose & Long edges remain insufficiently destroyed & --- \\
5 & Edit & Jointly modify destroyer and repairer & $\mathbf{9.088}$ \\
\bottomrule
\end{tabular}
\end{table}

This trajectory highlights two key behaviors enabled by the diagnostic mechanism: first, component attribution identifies the destroyer as the current bottleneck, prompting targeted modification; second, interaction analysis reveals coupling between operators, causing the search to transition from isolated component editing to joint design. This qualitative progression supports the system-level co-design results in the main text.

\subsection{Variation in Search Behavior}
\label{app:behavior}

Different random seeds can discover competitive designs through distinct search trajectories. To study this behavior directly, we additionally select one CVRP-DR instance and log the operation breakdown of three runs in Table~\ref{tab:actions}.

\begin{table}[!ht]
\centering
\footnotesize
\setlength{\tabcolsep}{4pt}
\caption{Search operation counts and outcomes on CVRP-DR across different random seeds.}
\label{tab:actions}
\begin{tabular}{@{}lccccc@{}}
\toprule
\textbf{Seed} & \textbf{Read} & \textbf{Edit} & \textbf{Diagnose} & \textbf{Evaluate} & \textbf{Best cost} \\
\midrule
2025 & $14$ & $122$ & $17$ & $81$ & $8.927$ \\
2026 & $11$ & $10$ & $163$ & $12$ & $\mathbf{8.281}$ \\
2027 & $12$ & $89$ & $102$ & $6$ & $9.138$ \\
\bottomrule
\end{tabular}
\end{table}

While the runs allocate effort differently, one skewing toward code modification, another toward diagnostic analysis, and the third mixing both, the diagnosis-intensive run yields the best final result in this sample. Although consistent with the diagnosis-guided search principle, a larger sample size would be required to establish a definitive correlation between operation distributions and final performance.

\subsection{Cross-Task Experience Transfer}
\label{app:transfer}

\begin{figure}[h]
\centering
\includegraphics[width=0.8\linewidth]{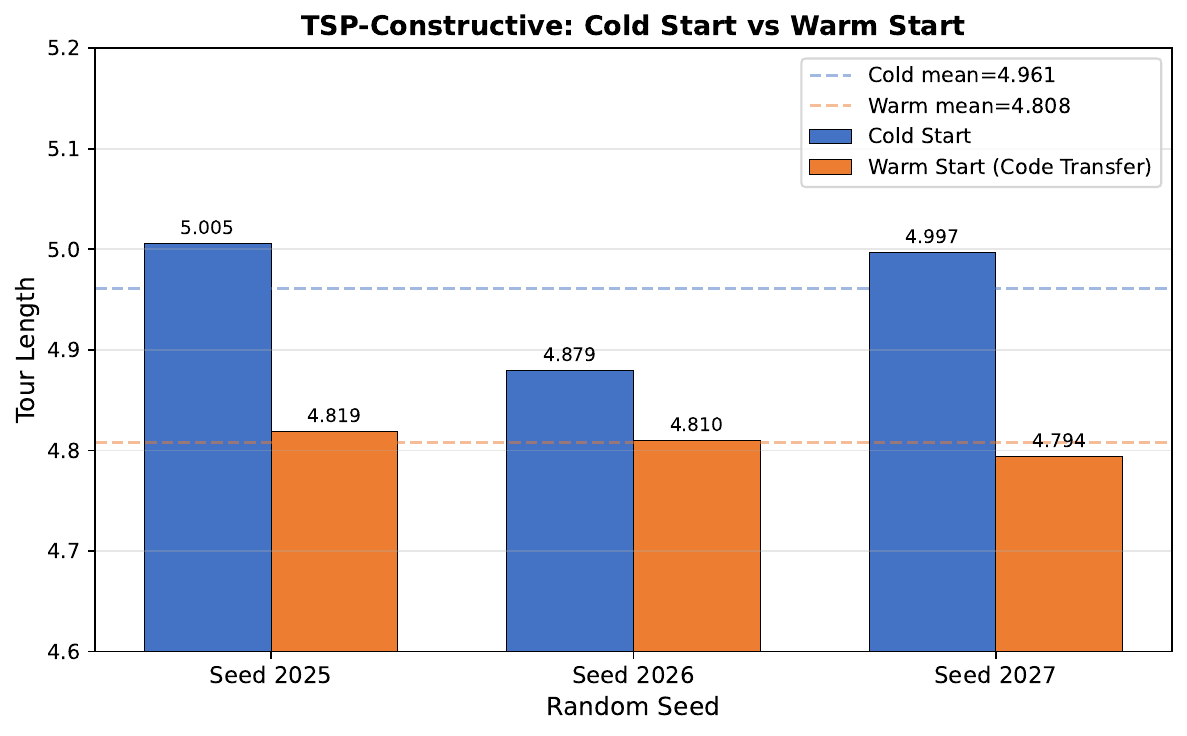}
\caption{Cold-start and warm-start performance on the plain TSP constructive task. Each group corresponds to one seed; dashed lines indicate group means.}
\label{fig:appwarm}
\end{figure}

Figure~\ref{fig:appwarm} presents a per-seed evaluation of the warm-start experiment. Initialized with a transferred design yielding a tour length of $5.017$, every warm-started seed outperforms its cold-started counterpart. Mean performance improves from $4.961$ to $4.808$, accompanied by reduced cross-seed variance.

This demonstrates that previously discovered algorithms serve as effective starting points for subsequent search rather than being discarded upon task completion, complementing the main findings on hierarchical experience accumulation. 

Note that this warm-start analysis uses the plain TSP constructive task (shipped nearest-neighbor baseline of $5.017$) rather than the standard TSP protocol used in the main evaluation (nearest-neighbour baseline of $6.824$); consequently, absolute values between these configurations should not be directly compared.

\subsection{Limitations and Future Work}
\label{sec:limitation}

Despite strong empirical performance across complex combinatorial optimization tasks, \systemname{} presents three limitations that motivate future work.

\textbf{Hand-Designed Situation Awareness.} Situations governing retrieval rely on discrete probes over a predefined category set. Consequently, these cues are hand-crafted rather than learned, meaning failure modes falling outside these categories may go undetected.

\textbf{Skill Overfitting.} Distilled skills derive from prior problem outcomes, exposing them to overfitting within aligned problem distributions and limiting zero-shot transfer to different domains. 

\textbf{Rigid Evaluation and Anti-Cheating Boundaries.} To guarantee evaluation integrity, our current protocol enforces strict execution harnesses, immutable entry points, and hard sandboxing boundaries. While this prevents reward hacking and private-pathway shortcuts, it can inadvertently suppress unconventional yet valid structural explorations that lie outside the predefined code contracts.

These limitations outline three promising avenues for future research.

\textbf{Learned Situation Awareness.} Replacing hand-designed categorical probes with a continuous, learned representation of the search state could enable more robust, adaptive perception.

\textbf{Optimized and Validated Skills.} Evolving skill refinement from posterior summarization to explicit optimization against held-out validation objectives, complete with rigorous versioning and testing before adoption, will actively mitigate overfitting.

\textbf{Flexible Yet Secure Validation Protocols.} Moving beyond rigid static sandboxes toward verifiable open-ended exploration represents a key frontier. Future systems could dynamically adapt safety and anti-cheating mechanisms to safely accommodate novel architectural designs, ensuring absolute integrity without stifling the creative boundaries of automated discovery.

\end{document}